\documentclass[pdflatex,sn-mathphys]{sn-jnl}

\jyear{2022}%

\theoremstyle{thmstyleone}%
\theoremstyle{thmstyletwo}%

\usepackage{rotating}
\usepackage{longtable} 
\usepackage{pdflscape}
\theoremstyle{thmstylethree}%

\begin{document}

\title[Kunhoth et al.]{Automated Systems For Diagnosis of Dysgraphia in Children: A Survey and Novel Framework}


\author*[1]{\fnm{Jayakanth } \sur{Kunhoth}}\email{j.kunhoth@qu.edu.qa}

\author[1]{\fnm{Somaya} \sur{Al-Maadeed}}
\equalcont{These authors contributed equally to this work.}

\author[1]{\fnm{Suchithra} \sur{Kunhoth}}
\equalcont{These authors contributed equally to this work.}

\author[1]{\fnm{Younus} \sur{Akbari}}
\equalcont{These authors contributed equally to this work.}

\affil*[1]{\orgdiv{Department of Computer Science and Engineering}, \orgname{Qatar University }, \orgaddress{\street{Al-Jamia Street}, \city{Doha}, \country{Qatar}}}


\abstract{Learning disabilities, which primarily interfere with the basic learning skills such as reading, writing and math, are known to affect around 10\% of children in the world. The poor motor skills and motor coordination as part of the neurodevelopmental disorder can become a causative factor for the difficulty in learning to write (dysgraphia), hindering the academic track of an individual. The signs and symptoms of dysgraphia include but are not limited to irregular handwriting, improper handling of writing medium, slow or labored writing, unusual hand position, etc. The widely accepted assessment criterion for all the types of learning disabilities is the examination performed by medical experts. The few available artificial intelligence- powered screening systems for dysgraphia relies on the distinctive features of handwriting from the corresponding images.This work presents a review of the existing automated dysgraphia diagnosis systems for children in the literature. The main focus of the work is to review artificial intelligence-based systems for dysgraphia diagnosis in children. This work discusses the data collection method, important handwriting features, machine learning algorithms employed in the literature for the diagnosis of dysgraphia. Apart from that, this article discusses some of the non-artificial intelligence-based automated systems also. Furthermore, this article discusses the drawbacks of existing systems and proposes a novel framework for dysgraphia diagnosis.}

\keywords{Dysgraphia diagnosis, Handwriting disability, Machine Learning , Automated Systems}



\maketitle

\section{Introduction}\label{sec1}

\begin{figure}[h]%
\centering
\includegraphics[width=0.9\textwidth]{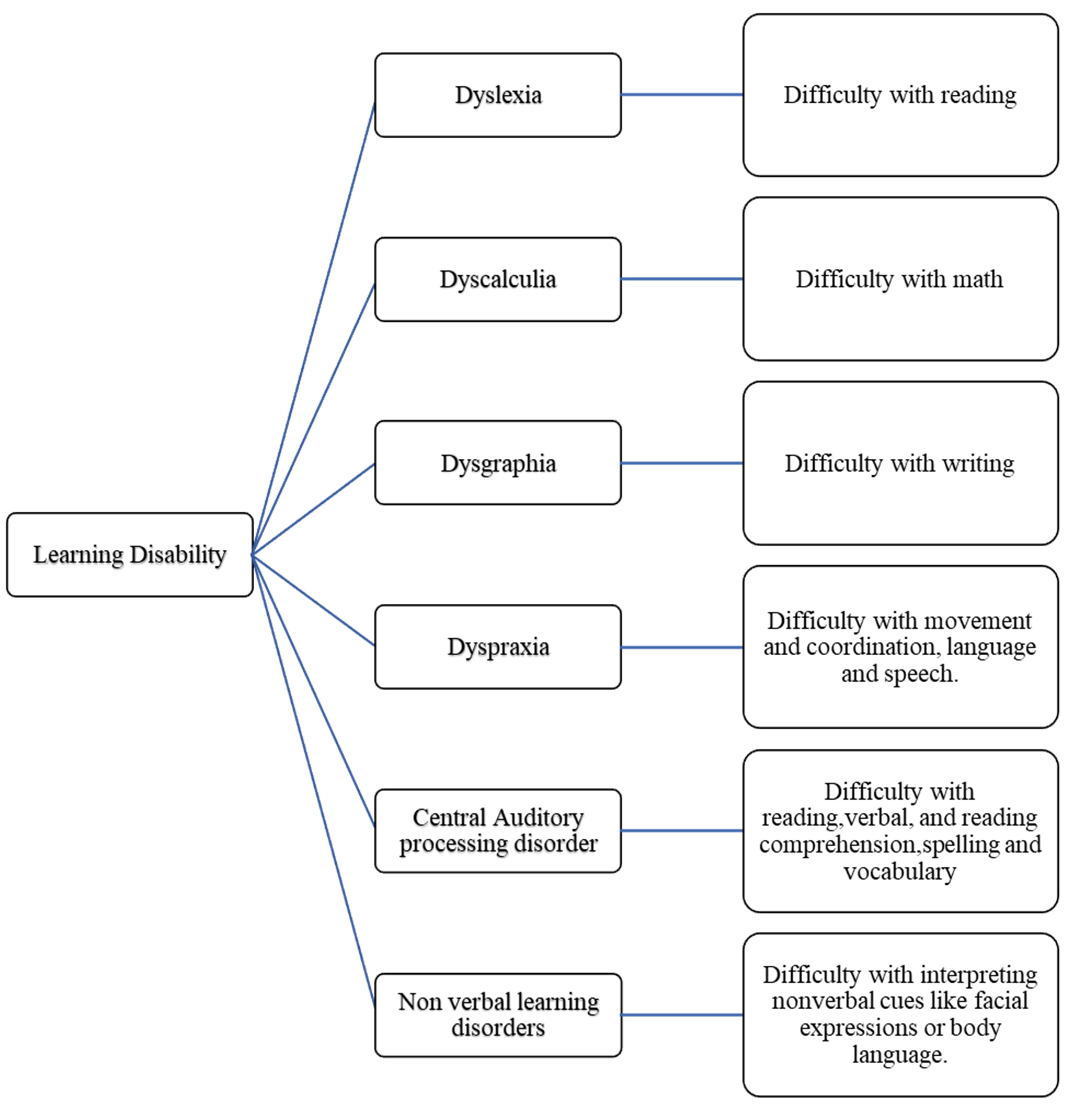}
\caption{Types Of Learning Disabilities}\label{fig1}
\end{figure}

Learning disabilities or learning disorders, a hypernym for a wide variety of learning problems hinders the skill acquisition activity of an individual. It is not a problem with intelligence. Nevertheless, it can have a negative impact on the self-esteem and confidence of children who are expected to acquire new information and skills day by day. Students with learning disabilities account for the major proportion of “special educational needs” category \cite{Knickenberg2020}. Since these disabilities affect the perception capability of a child, the difficulties may be either in reading, writing, doing math, or any other tasks. Figure \ref{fig1} shows different common learning disabilities found in children.  

\begin{figure}[h]%
\centering
\includegraphics[width=0.9\textwidth]{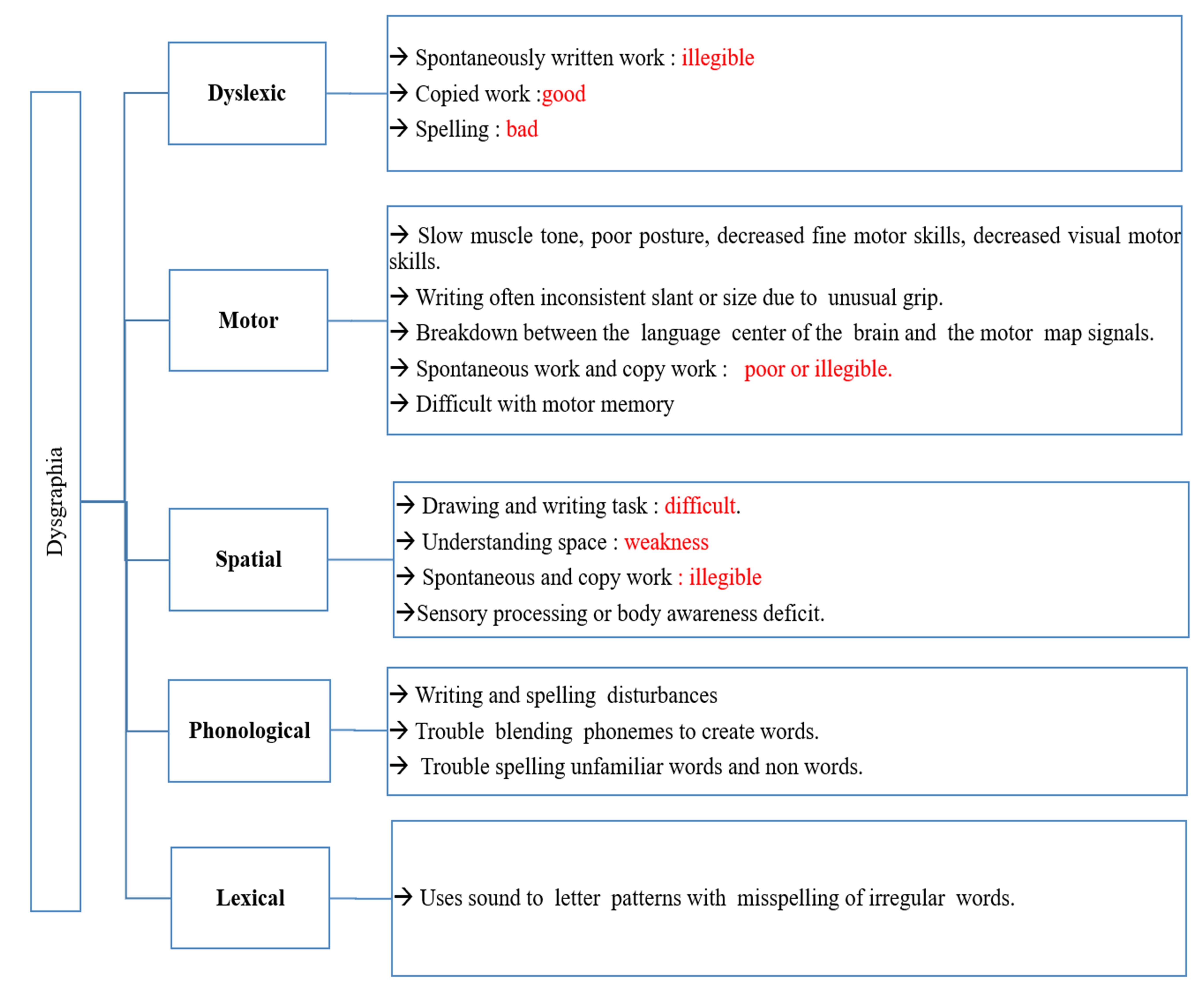}
\caption{Types Of Dysgraphia}\label{fig2}
\end{figure}

Dyslexia \cite{Lyon1995} is a kind of learning disability that includes difficulty in reading because of issues in recognizing the speech sounds as well as decoding them. Dyslexia is affected in the brain areas which is assigned for processing the language. Dyscalculia \cite{Keong2020} is a specific type of learning disorder that involves difficulty in understanding numbers and its related issues in learning mathematics. Dyspraxia \cite{Gibbs2007} a.k.a developmental coordination disorder affects the coordination skills of individuals which hinders them from accomplishing tasks such as playing sports, driving, etc that requires balance.  Central auditory processing disorder (CAPD) \cite{Koravand2017} involves difficulty in hearing in children. This is because of the lack of coordination between the brain and ears. Children with CAPD have trouble in understanding when they hear sounds.  Dysgraphia is primarily considered as a disorder in written expression. It can affect the spelling, grammar, organization, etc.,  in addition to the handwriting aspects \cite{Deuel1995}. Figure \ref{fig2} summarizes the basic categorization of dysgraphia based on the related symptoms . Although the exact prevalence depends on the definition of dysgraphia, between 10-30 \% of children face difficulty in handwriting \cite{Chung2020}.

The typical diagnostic strategy for any learning disability follows a team-based assessment \cite{Chung2020}, which involves the support of an occupational therapist, speech therapist, special education teacher, psychologist etc. In addition, any other prevailing medical conditions such as poor vision, hearing problems, intellectual disability, lack of proper training should be ruled out with the help of a specialist.  As far as dysgraphia is concerned, it is indeed important to consider the several contributing factors such as handwriting speed and legibility, inconsistency between spelling ability and verbal intelligence quotient, as well as the pencil grip and writing posture evaluation. However, there is no generalized medical testing strategy available for the diagnosis of dysgraphia. Concise Evaluation Scale for Children’s Handwriting (BHK) for French \cite{LHamstra-BletzdeBieJ1987}, Detailed Assessment of Speed of Handwriting (DASH) in Latin \cite{Barnett2009} are some commonly used standards in the assessment of dysgraphia.

\begin{figure}[h]%
\centering
\includegraphics[width=0.9\textwidth]{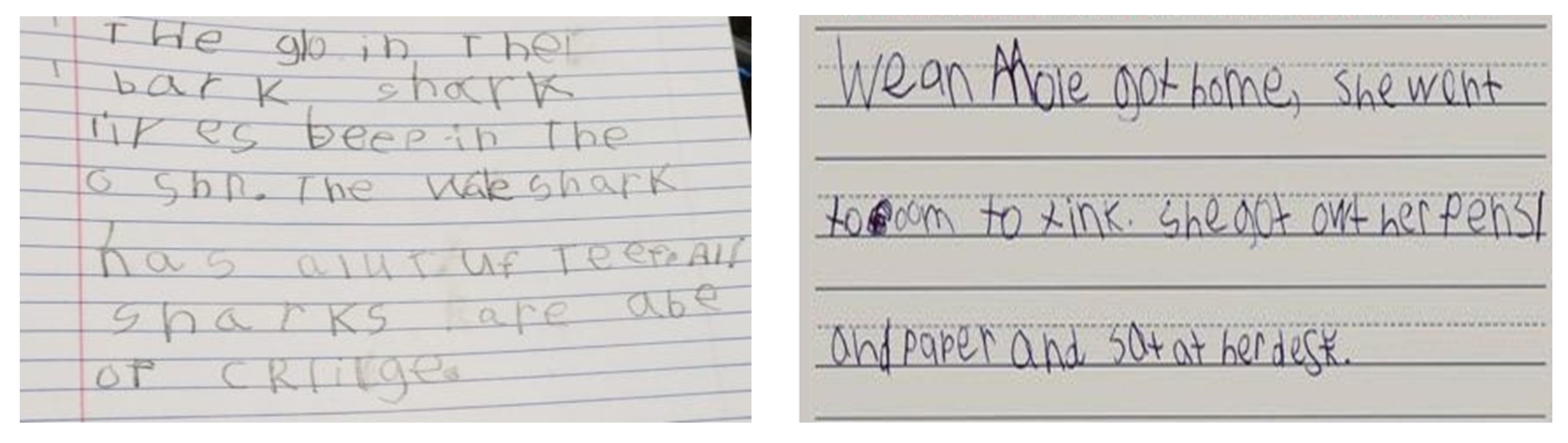}
\caption{Handwriting samples of person with dysgraphia }\label{fig10}
\end{figure}

It is quite difficult to diagnose any of the learning disorders because of the multiple cues that should be considered in the assessment. Depending on the age and developmental stage, the warning signs and symptoms may vary too. More importantly, the predefined symptoms should persist for at least 6 months with parallel intervention measures being administered \cite{AmericanPsychiatricAssociation.2013}. Dysgraphia may appear in isolation or as a comorbidity with other learning disorders or even autism spectrum disorder, developmental coordination disorder (DCD) \cite{Lopez2018}, and Attention Deficit Hyperactivity Disorder. This 
20
 emphasizes the timely diagnosis of dysgraphia or handwriting disorders in particular. The early recognition and intervention lessen the task and efforts needed to correct the disorders. 
The manual assessment techniques solely rely on the handwritten product for the final scoring and judgment. This paved the way to propose several automated techniques that can exploit the dynamic characteristics of handwriting as well. Digital tablets capable of capturing these multiple features of handwriting have yielded promising results in the related research. Meantime studies indicate that the writing sensation may be different for stylus-tablet setting in contrast to pencil-paper, which is the commonly used procedure during the skill acquisition phase \cite{Gargot2020}. Despite the complexity and the multifaceted approaches involved in the diagnosis of dysgraphia, several artificial intelligence-based and non artificial intelligence based automated  technologies have been proposed in recent years\cite{rosenblum2018inter,vaivre2021phenotyping,chang2013handwriting}. 

To the best of our knowledge, no other work in the literature has focused on the review of automated dysgraphia diagnosis systems until now. There are a few related review articles in the literature and their details are tabulated in table 1. Most of the review articles in the literature targeted all learning disabilities instead of one. And few of them have not addressed dysgraphia at all.  Since the trend of automated systems for dysgraphia diagnosis is increasing in the last few years, there is a need for a review article on the same. It can help researchers and academic students to learn more about this topic. 

This work presents a review of the existing automated dysgraphia diagnosis systems for children in the literature. The main focus of the work is to review artificial intelligence-based systems for dysgraphia diagnosis in children. This work discusses the data collection method, important handwriting features, machine learning algorithms employed in the literature for the diagnosis of dysgraphia. Apart from that, this article discusses some of the non-artificial intelligence-based automated systems and psychological methods also. Furthermore, this article discusses the drawbacks of existing systems and proposes a novel framework for dysgraphia diagnosis. 

\begin{table}[]
\caption{Summary of related works}
\begin{tabular}{|l|p{8cm}|}
\hline
Reference  & Article Summary                                                                                              \\ \hline
Vanitha and Kasthuri\cite{vanithadyslexia}    & Review of machine learning algorithms used in Dyslexia prediction                                                              \\ \hline
Chakraborty et al.\cite{chakraborty2019survey}    & It presents a survey paper  on the topic machine learning algorithms for learning disability prediction. But the presentation was some what abstract level and no details about dysgraphia diagnosis systems                                          \\ \hline
Vanjari et al.\cite{vanjari2019review}   & A review on learning disabilities and technologies determining the severity of learning disabilities               \\ \hline
Prabha and Bhargavi\cite{jothi2019prediction}   & A similar work like \cite{vanithadyslexia} an breif review on prediction of dyslexia using machine learning.                                                \\ \hline
Saxena and Saxena\cite{saxena2020machine}    & Reviews and explains the role of machine learning in learning disabilities diagnosis but lacks the insight in to dysgraphia.                                                         \\ \hline
Jamhar et al. \cite{jamhar2019prediction}   & A systematic review about machine learning methods for learning disorder prediction.But not addressed about dysgraphia. Further more the focus was just on the machine learning algortihms not about data collection or features.                                                               \\ \hline
\end{tabular}
\end{table}

\section{Research Methodology}\label{sec2}

This work followed a systematic approach for preparing the literature review. The systematic approach is followed to find out the specific issues in this research domain.  In the systematic approach, we defined certain search terms and pre-selected a few digital databases to find the related research articles. After finding the related articles  the most relevant works are sorted out and reviewed in this work.  

A research question is a required element for conducting the review in a systematic approach. The generic research questions in this work are:

\begin{itemize}

\item	What are the automated systems used for the prediction of dysgraphia in children?
\item	How effectively the machine learning technology has employed in the literature for the diagnosis of dysgraphia in children?
\item	How the handwritten data are collected for training machine learning models and what are the relevant handwritten features which can discriminate the abnormal and normal handwriting. 
\end{itemize}

During the literature search, we utilized four popular digital libraries IEEE Digital Library, Web of Science, PubMed, and Springer Link to find the related articles. The search keywords are used for literature search are given below. 

\begin{itemize}

\item	Automated System Dysgraphia 
\item	Machine Learning Dysgraphia
\item	Deep Learning Dysgraphia 
\item	Automated System Learning Disorder
\item	Machine Learning for Learning  Disorder

\end{itemize}

Among all the obtained search results by using the above-mentioned keywords separately in each database, we found that only about 50 papers are most relevant to the specific topic. Another interesting fact is that most of the papers in the literature are published after 2015. 

\section{Dysgraphia Diagnosis Methods}\label{sec3}
In this section, we first discuss the psychological methods used to diagnose dysgraphia and the manual analysis of handwriting. We then review the notable contributions to automated analysis (machine learning methods) and non-ML based diagnosis systems of handwriting
 for the detection of dysgraphia.
 
\subsection{Psychological Methods}\label{sec3-1}
Dysgraphia is a Greek word with the meaning, difficulty, or poor (Dys) writing  (graph). Initially, the difficulty in handwriting or dysgraphia was described by Hamstra-Bletz and Bolte as difficulty in the construction of letters while writing which is closely related to the mechanics of writing \cite{hamstra1993longitudinal}. It has also been referred to as a specific learning disability \cite{brown1981learning,rosenblum2004handwriting}.

\begin{figure}[h]%
\centering
\includegraphics[width=0.9\textwidth]{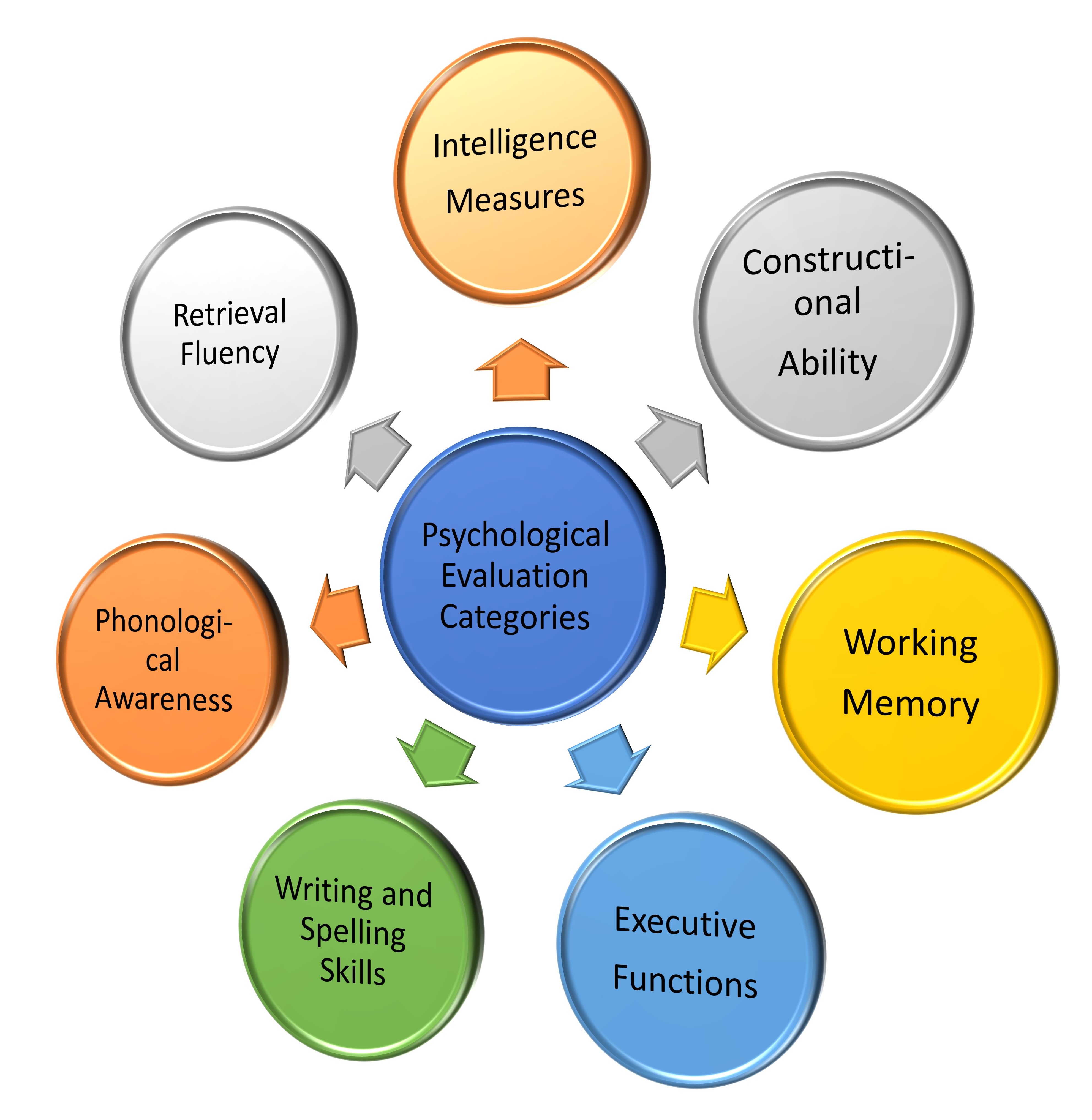}
\caption{Psychological evaluation: Assessment categories   }\label{fig0}
\end{figure}

The conventional approaches for dysgraphia diagnosis consist of two sections: academic skills examination and cognitive skills analysis. The popular symptoms of dysgraphia include “messy handwriting, inconsistency in letter spacing and capitalization, pain or discomfort when writing, fine motor skill challenges, trouble with spelling, or trouble with composing written work”. And further, it is also common that the students with dysgraphia can express themselves well while speaking but can’t transfer them onto paper perfectly. The occupational therapist or trained psychologist evaluate different skills (falling under different category (figure \ref{fig0})) of the student to find out the existence of dysgraphia.

The constructional ability assessment will look for the proficiency of the student to copy or reconstruct lines, shapes, or figures. One of the popular conventional methods utilized by trained psychologists for constructional ability assessment is "Beery Visual Motor Test of Integration – Sixth Edition (VMI-6)” \cite{beery2004beery}. In the VMI-6, the participants are asked to copy drawings on paper. The complexity of the drawings will be increased throughout each stage of the test and participants are not allowed to erase the drawings. The drawings given for the copying task include overlapped figures, angles, and three-dimensional images. But the visual motor skills purely depend on the age and it is expected that until age 16, the standards of the visual-motor skills will be different for every age until 16. Usually, many students with dysgraphia struggle with eye-hand coordination and planning. The visual-motor integration test conveys more information about the participant’s capacity to understand a drawing and the motor ability to copy the information (motor response). Other popular constructional ability assessment tests are Bender Gestalt II \cite{mehrinejad2012investigation}, NEPSY-II \cite{korkman2014nepsy}.

Executive function skills enable the student to plan, concentrate, remember commands and organize multiple tasks. The student requires all the executive functions to intact for writing \cite{chung2015dysgraphia}. Rey Complex Figure Test \cite{meyers1995rey} and  Behaviour Rating Inventory of Executive Function (BRIEF) \cite{roth2014assessment} are the popular examination test conducted for executive function assessment. In Rey Complex Figure Test, the students were asked to draw very complex figures. Based on their drawing output the capacity of executive functions is quantified. On the other hand, BRIEF provides a questionnaire to students’ parent and teacher, and ask them to answer for that. The provided questionnaire is a rating form that consists of 86 items to assess eight clinical scales independently. The eight clinical scales include three behavioral regulation scales (Inhibit, shift, emotional control)and five metacognition scales (initiate, working memory, plan, monitor, organization of materials ). 

Writing and spelling skills, phonological awareness, retrieval fluency are commonly found to be less in children with dysgraphia \cite{mccloskey2017developmental}. Wechsler Individual Achievement Test (WIAT-III)\cite{burns2010wechsler}, Woodcock Johnson-III Tests of Achievement (WJ-III) \cite{woodcock2007woodcock}, and Test of Written Language-4 (TOWL-4)\cite{hammill2009test} can assess the writing and spelling skills. Phonological awareness refers to the ability of the human to perceive and work with the audio, especially the sound in the spoken language. It includes understanding patterns like alliteration or rhymes, the ability to segment the sentences into words, understanding of phonemes and syllables, etc. Comprehensive Test of Phonological Processing\cite{wagner1999comprehensive}, NEPSYII phonological processing are the two popular phonological awareness assessment tests.

Working memory refers to the cognitive ability of the human memory system that has limited capacity and can keep information for short time. Generally, dysgraphia can occur with some degree of working memory problems\cite{crouch2007dysgraphia}. Test of Memory and Learning – 2 (TOMAL-2) \cite{reynolds2007test} and Wide Range Assessment of Memory and Learning-2 (WRAML-2)\cite{hartman2007wide} are two popular examinations for assessing the decline of working memory. Intelligence measure a.k.a intelligence quotient quantifies human intelligence which varies with the age.  Wechsler Intelligence Scale for Children (WISC-IV)\cite{petermann2011wechsler}  and Differential Ability Scales (DAS)\cite{elliott1990differential} are the popular tests for quantifying the intelligence of children.

The Concise Evaluation Scale for Children’s Handwriting (BHK) is another popular manual handwriting analysis method conducted under the supervision of an occupational therapist or psychologist to quantify the speed and standard of the writing. 
BHK test is usually conducted in the individual clinical setting and classical scholar setting. Initially, it was introduced for assessing the handwritten samples of 2nd and 3rd-grade students. Currently, BHK scales are used in the research for constructing the ground truth of the data used for training and evaluation of machine learning-based dysgraphia diagnosis systems.

\subsection{Machine Learning  Based Methods}\label{sec3-2}

\begin{figure}[h]%
\centering
\includegraphics[width=0.9\textwidth]{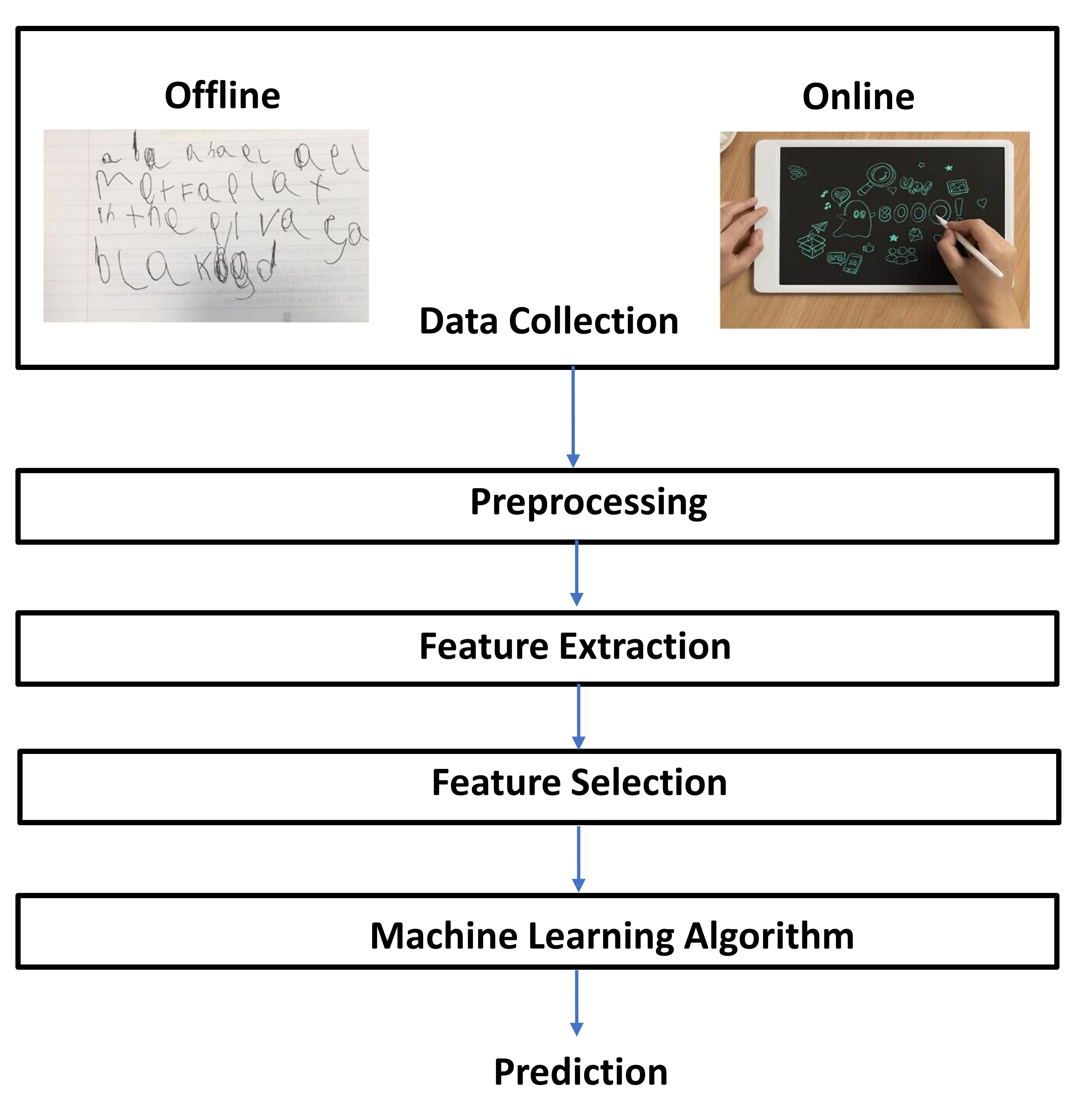}
\caption{General work flow of machine learning based dysgraphia diagnosis system}\label{fig3}
\end{figure}

These section discusses machine learning based systems for diagnosis of dysgraphia from either the images of handwritten texts or features of writing dynamics. The general flow of machine learning based dysgraphia diagnosis system is shown in figure \ref{fig3} . As like in other machine learning applications,  dysgraphia diagnosis systems also follow similar workflow. Data collection, preprocessing, feature extraction, feature selection, and training using machine learning classifier are important steps included in building the machine learning-based dysgraphia diagnosis system. Data is a pivot for any machine learning task. Cleaned and sufficient data is required for any machine learning algorithms to make the accurate prediction. In the case of the dysgraphia diagnosis system, two types of data collection methods are usually implemented. The first approach is offline-based data collection where the subject is asked to write or copy a few words or sentences on the paper or tablet and the resulting handwritten images are used for further analysis. On the other hand, the online-based data collection is focused on collecting the handwritting data during the run time which includes the trajectory of the pen /pencil, writing speed, pressure on pen tip, etc.  Most of the existing dysgraphia diagnosis systems followed the online data collection strategy. 

\subsubsection{Feature Extraction}\label{subsec2}

\begin{figure}[h]%
\centering
\includegraphics[width=0.9\textwidth]{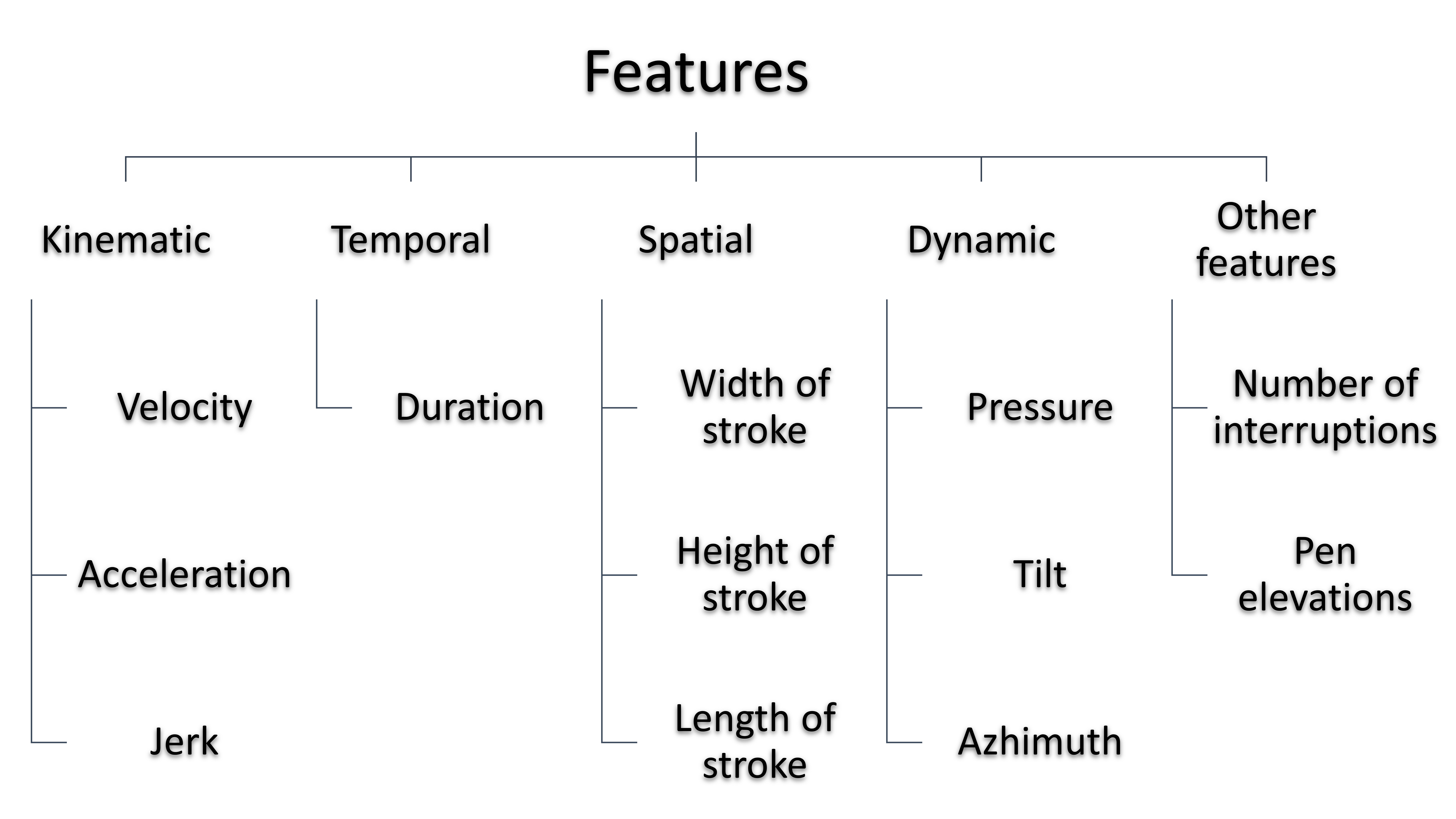}
\caption{Relevant handwriting features}\label{fig5}
\end{figure}

Extraction of features from the data is crucial for training the machine learning models. Multiple features are extracted from the collected handwritten data for further analysis in the existing systems. The important online handwritten features used for machine learning-based systems in the literature for dysgraphia detection are shown in figure \ref{fig5}. 

\begin{itemize}

\item Kinematic Features
\end{itemize}

The kinematics features mainly include the velocity,  acceleration, and jerk of the writing. Velocity just quantifies the speed of writing and abnormal variability in writing speed is related to the underlying handwriting problems. Acceleration computes the variability in the velocity during writing. Jerk is the change in acceleration over time per stroke. Unusual changes in acceleration over time per stroke may be related to writing problems. 
Various studies \cite{kushki2011changes,rosenblum2003air} in the literature have found that the kinematic aspects are affected in children with handwriting disabilities.Along with the average velocity of the whole writing process, few works considered the velocity or speed of writing each word or letter as a new derived feature for enhanced discrimination.   

\begin{itemize}

\item Temporal Features
\end{itemize}

The temporal features mainly quantify the writing/drawing duration or time. Usually, the children with dysgraphia will take more time to write or draw compared to normal children. In this sense, the writing or drawing duration has significance in differentiating the normal and abnormal handwriting behavior. The total time required to complete the task, the total time spent on the paper, the ratio between the total time spent on the paper and total time of the task, etc are the popular temporal or time-related features utilized in the literature for dysgraphia detection. 
\begin{itemize}

\item Spatial Features
\end{itemize}

The width, height, and length of the strokes as well as whole written data are the common spatial features extracted from handwritten data.  The writing or drawing activities of children with dysgraphia always contain various inconsistencies. Irregular size of letters, irregular spacing between the words are the common inconsistencies found in the writing. The spatial feature can well discriminate the writing samples of  children with dysgraphia.This features can be called as geometric features also. The offline handwritten based dysgraphia diagnosis systems mainly rely on these features to differentiate the normal and abnormal hand writings . 
\begin{itemize}

\item Dynamic Features
\end{itemize}

The popular dynamic features are pressure, tilt, azimuth, etc. Among these, pressure features are quantified to find out the characteristics of pressure induced by the pen tip on the writing surface. Generally, the statistical measures such as mean, median, the standard deviation of pressure are quantified for feature construction. Speed of pressure change, speed of pressure change frequencies are the other derivable features from pressure values. The tilt feature is used to measure the inclination of the pen or pencil used for writing. The azimuth angle of pen or pencil with respect to the plane of written surface is also quantified for feature set construction.

\begin{figure}[h]%
\centering
\includegraphics[width=0.9\textwidth]{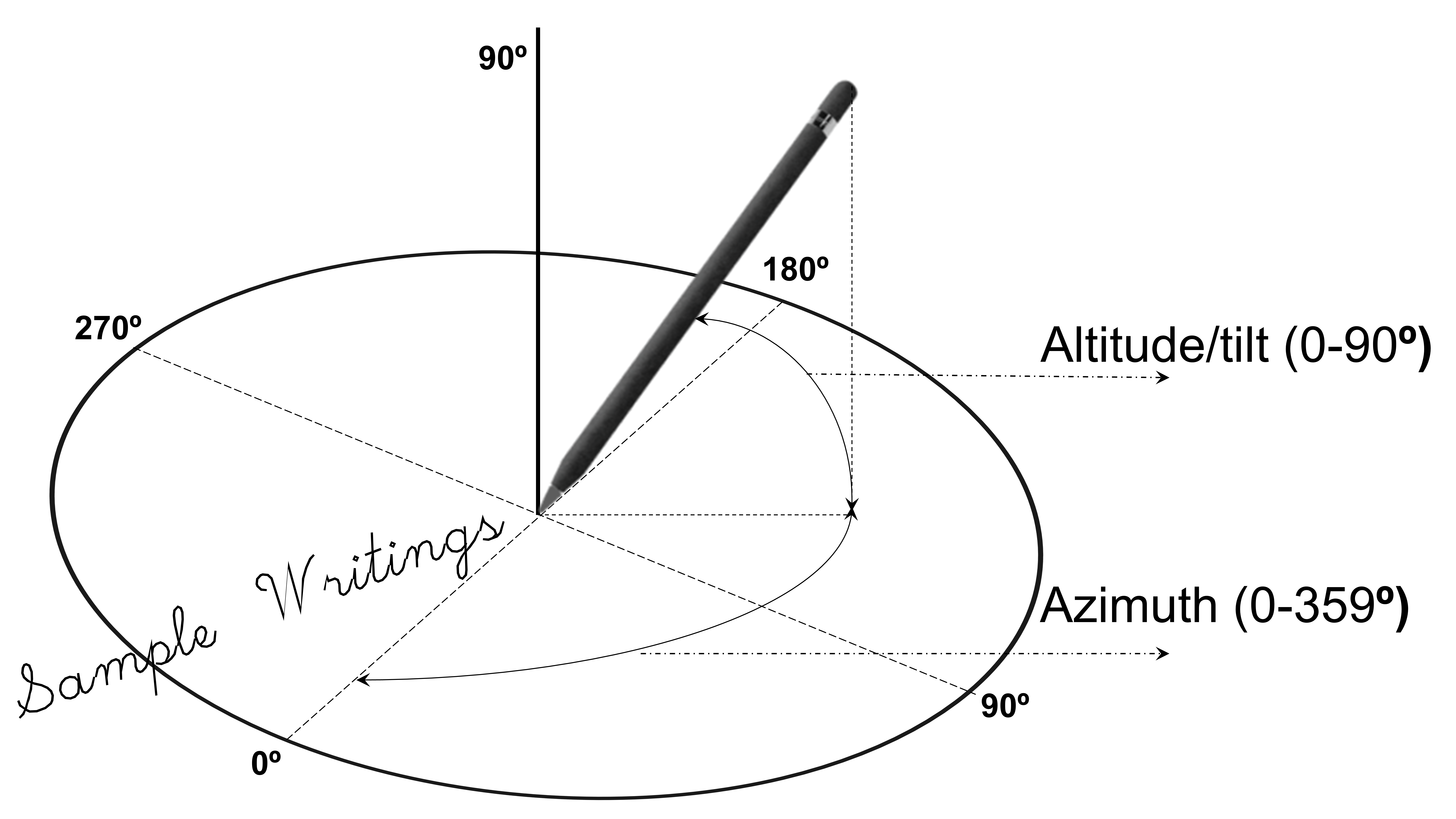}
\caption{Tilt and Azimuth angle }\label{fig6}
\end{figure}
\begin{itemize}

\item Other Features
\end{itemize}

Apart from the main four categories of features ( Kinematic , temporal , spatial ,dynamic), few works in the literature have considered some other features to improve the dysgraphia screening efficiency . It includes number of interruptions occurred during the writing , number of pen elevations , as well as number of times the child has made the mistake and erased it ( erase count ) etc. These features has also have relevance in screening dysgraphia , since children with dysgraphia are tends to make more mistakes than normal children while writing. 

Most of the works in the literature has used combination of different feature categories to effectively identify the existence of dysgraphia in children using machine learning algorithms. Table \ref{tab:3} displays the different feature combinations used in the literature .

\begin{table}[h]
\caption{ Summary of feature combinations used in the literature}
\begin{tabular}{|l|lllll|}

\hline
\multirow{2}{*}{References} & \multicolumn{5}{c|}{Features}                                                                                                                                                                                                 \\ \cline{2-6} 
                            & \multicolumn{1}{l|}{Kinematic}                 & \multicolumn{1}{l|}{Temporal}                  & \multicolumn{1}{l|}{Spatial}                   & \multicolumn{1}{l|}{Dynamic}                   & Other                     \\ \hline
                           Mekyska et al.\cite{Mekyska2017} & \multicolumn{1}{l|}\checkmark & \multicolumn{1}{l|}- & \multicolumn{1}{l|}- & \multicolumn{1}{l|}\checkmark &\checkmark\\ \hline
                            Asselborn et al. \cite{Asselborn2018}& \multicolumn{1}{l|}\checkmark & \multicolumn{1}{l|}\checkmark & \multicolumn{1}{l|}\checkmark & \multicolumn{1}{l|}\checkmark & - \\ \hline
                           
                           Gargot et al. \cite{Gargot2020} & \multicolumn{1}{l|}\checkmark & \multicolumn{1}{l|}\checkmark & \multicolumn{1}{l|}\checkmark & \multicolumn{1}{l|}\checkmark & - \\ \hline
                           
                           Drotar et al.\cite{Drotar2020} & \multicolumn{1}{l|}\checkmark & \multicolumn{1}{l|}\checkmark & \multicolumn{1}{l|}\checkmark & \multicolumn{1}{l|}\checkmark & \checkmark \\ \hline
                           Asselborn et al. \cite{Asselborn2020}  & \multicolumn{1}{l|}\checkmark & \multicolumn{1}{l|}\checkmark & \multicolumn{1}{l|}\checkmark & \multicolumn{1}{l|}\checkmark & - \\ \hline
                           
                            Devillaine et al.\cite{Devillaine2021}& \multicolumn{1}{l|}\checkmark & \multicolumn{1}{l|}-& \multicolumn{1}{l|}\checkmark & \multicolumn{1}{l|}\checkmark & \checkmark\\ \hline
                            Deschamps et al. \cite{Deschamps2021} & \multicolumn{1}{l|}\checkmark & \multicolumn{1}{l|}\checkmark & \multicolumn{1}{l|}\checkmark & \multicolumn{1}{l|}\checkmark & - \\ \hline
                           
                           Dankovicova et al.\cite{Dankovicova2019} & \multicolumn{1}{l|}\checkmark & \multicolumn{1}{l|}\checkmark & \multicolumn{1}{l|}- & \multicolumn{1}{l|}\checkmark & \checkmark \\ \hline
                           
                           Rosenblum et al. \cite{rosenblum2016identifying} & \multicolumn{1}{l|}- & \multicolumn{1}{l|}\checkmark & \multicolumn{1}{l|}\checkmark & \multicolumn{1}{l|}\checkmark & - \\ \hline
                           Mekyska et al. \cite{Mekyska2019} & \multicolumn{1}{l|}\checkmark & \multicolumn{1}{l|}\checkmark & \multicolumn{1}{l|}\checkmark & \multicolumn{1}{l|}\checkmark & \checkmark \\ \hline
                           Devi et al. \cite{devi2021early} & \multicolumn{1}{l|}\checkmark & \multicolumn{1}{l|}\checkmark & \multicolumn{1}{l|}- & \multicolumn{1}{l|}\checkmark & - \\ \hline
                            Kedar \cite{kedar2021identifying}& \multicolumn{1}{l|}\checkmark & \multicolumn{1}{l|}\checkmark & \multicolumn{1}{l|}\checkmark & \multicolumn{1}{l|}\checkmark & - \\ \hline
\end{tabular}
\label{tab:3}
\end{table}

\subsection{Existing Machine Learning Based Systems}\label{subsec2}

Reference \cite{Mekyska2017} proposed a method for automated diagnosis of developmental dysgraphia and estimation of handwriting difficulty level using handwriting analysis. A digitized tablet is used to acquire the handwriting data from the students.  A total of 54 students ( 27 normal and 27 dysgraphic ) are selected for experiments. The students are made to write seven semi Hebrew letters on A4-sized paper attached to the surface of WACOM Intuos II xy digitizing tablet (404 × 306 × 10 mm) using a wireless electronic pen with a pressure-sensitive tip (Model GP-110). The Computerized Penmanship Evaluation Tool (COMPET) is used for collecting the data. COMPET is a popular and standardized tool for online data acquisition and analysis. From the acquired handwriting data, three different kinds of features: ten Kinematic features, 34 non-linear dynamic features, and  7 other features are extracted for training artificial intelligence models for the classification of dysgraphic students. Random forest algorithm \cite{Biau2012} and Linear discriminant analysis algorithm \cite{Izenman2013} were used to train the machine learning models. The obtained results indicated that the Random forest classifier can classify the dysgraphic and normal subjects with 96\% sensitivity and specificity. Furthermore, the finding indicated that the altitude/tilt and pressure can discriminate the subjects well.

An automated dysgraphia diagnosis tool for primary school students using the consumer-level tablet was proposed in \cite{Asselborn2018}.  In the proposed work, 298 students including 56 dysgraphic individuals were instructed to write on a sheet of paper affixed to a Wacom Intous tablet for 5 mins. Ductus software ( LPNC laboratory ) tool is used for data collection when the students are writing. From the handwriting task, 54 features that define various characteristics of the handwriting are extracted for further analysis. The handwriting features include Static features: purely geometric characteristics of a written text, Kinematic features: dynamics of handwriting path, Dynamics features: characteristics of the pressure recorded between the pen tip and the tablet surface, characteristics of the pen tilt. The extracted features are utilized to train a machine learning classifier, Random Forest, for intelligent decision-making. The proposed machine learning model has achieved excellent accuracy for diagnosing dysgraphia. 

A study was conducted on 76 elementary school students in the Czech republic for a computerized assessment of graph motor difficulties or dysgraphia \cite{Mekyska2019}. The cohort of students was asked to draw 7 different figures such as the Archimedean spiral, connected loops , rainbow etc on an A4 paper, that was laid down and fixed to a digitizing tablet Wacom Intuos Pro L (PHT-80). Five different categories of features were extracted from the acquired handwriting data. The extracted features include spatial: width, height, and length of the whole product, as well as its strokes, stroke width, height, and length, temporal: duration of drawing, kinematic – velocity, acceleration, and jerk, dynamic: pressure, tilt, and azimuth,  other – number of interruptions (pen elevations and relative number of interruptions). Mann-Whitney U test \cite{LaerdStatistics2015} was conducted to compute the very relevant features among extracted features. An ensemble machine learning algorithm XGBoost \cite{Chen2016} is trained with the extracted features for automated detection of dysgraphia in children. 

A tablet-based smart application was developed for early prediction of the potential risk of handwriting alteration in children during the preliteracy stage \cite{Dui2020} . The main aim of the proposed approach was to develop a tool to anticipate dysgraphia screening in preschoolers. For the data collection, 104 preschoolers were asked to use an application named ‘ Play-Draw-Write’ on the tablet device to draw basic shapes such as circles, squares, lines, etc. in different sizes using a stylus pen. Gesture smoothness, pressure, drawing kinematics features collected during writing and drawing task are collected for further analysis. The logistic regression \cite{Peng2002}]model is trained using these features for the classification task. Even though the proposed approach can classify the handwriting alteration it is not possible to confirm that real handwriting difficulties would arise in the following year schooling stage.

In \cite{Gargot2020} 280 school-going children ( average age about 9 years) are selected for conducting the BHK test on digital tablets. Twelve different features of handwriting tasks including the static, kinematic, and dynamic features are extracted for further analysis. An unsupervised method named K-means clustering  is employed to distinguish the students with dysgraphia based on the severity levels ( mild or severe ). Linear regression models are used to predict handwritten quality scores. 

A machine learning-based system for dysgraphia is proposed in \cite{Drotar2020}. The proposed approach focused on collecting a new dataset as well as analysis of the data to detect the existence of dysgraphia disability in school children. 120 school students have participated in their study for building the handwritten data set. Trained professionals from the center for children with special needs are employed for data collection from dysgraphic children. The participated children were asked to write specific letters, words, and sentence on the paper attached to the screen of the WACOM Intuos Pro Large tablet using the normal pen. The tablet can capture five different signals: pen movement in the x-direction, pen movement in the y-direction, the pressure of the pen on the tablet surface, and the azimuth and altitude of the pen during handwriting. 22 types of features including spatiotemporal and kinematic features were extracted from the collected data. Multiple machine learning algorithms were utilized in this work for classification and their performances are compared.  The obtained results indicate that the AdaBoost algorithm \cite{Zizka2019} achieved the highest classification accuracy of  80\%. Among all the extracted features, pressure and pen lifts are the features with high discriminatory potential.

Images of handwritten text alone were used \cite{Richard2020} for detecting students with writing disabilities. The proposed work is based on the fact that in the handwritten task the dysgraphic children will face difficulties in the various process such as the formation of the letter, maintaining the size of letters, following the straight line, maintaining the consistent gap between letters as well as words. About 1400 handwritten images where each image contains four lines of handwritten text. From the images, different features including slant, pressure, amplitude, letter spacing, word spacing, slant regularity, size regularity, size regularity, horizontal regularity were manually extracted and four different machine learning classifiers are trained for dysgraphia prediction. Among four classifiers the Random forest classifier achieved the best accuracy of 96.2

Reference \cite{Asselborn2020} proposed various scales to evaluate handwriting difficulties in a modernized way. In the proposed work, the participated 448 children were asked to write five sentences used in the BHK test using an application deployed in the IPad device. Dynamico software is used for collecting handwritten raw data. From the raw data, for each student 63 features falling under four categories: static, pressure, kinematic, tilt are extracted. Principle component analysis is employed for projecting the 63 features to three-dimensional feature space. An unsupervised machine learning algorithm K-Means clustering is employed for grouping the normal and dysgraphic subjects to different clusters. 

The raw data obtained from graphomotor tests are analyzed using machine learning algorithms for the early detection of dysgraphia in children\cite{Devillaine2021}. Initially, 305 students were asked to participate in two different sets of experiments: the BHK test and the graph motor test. The BHK test is performed to label the students as dysgraphic or not for training the machine learning models. No features or data from the BHK test are used for machine learning-based analysis. After labeling, the students are asked to perform a graphomotor test, where each student has to write or draw specific predefined shapes or figures called stimuli on the paper attached to the screen of the tablet. Numerous features including time duration of the stimulus, duration of a stroke , duration of lift, velocity, jerk , Renyi Entropy of order 2, signal to noise ratio, etc were extracted from the raw data for further analysis. Among participating students, only 43 children were having dysgraphia, so the whole dataset ( raw data from graphomotor task) are divided into two: Active database and Z-score database. The active database contains the data of 43 children with dysgraphia and 43 other children without dysgraphia. The data of the remaining 219 non-dysgraphic children were added to the Z-score database. The data in the Z-score database is not used for training the machine learning model, instead, it is used for correcting the features in the active database. After correcting the features in the active database, relevant features are selected using Linear SVM and Extra Trees algorithm and utilized for training machine learning algorithms. Various machine learning algorithms including SVM \cite{Burges1998}, Random Forest, MLP\cite{Yegnanarayana1994}, Extra Trees, Gaussian Naïve Bayes\cite{Wu2008}, Ada Boost, etc are utilized in the proposed work for analyzing the features.  Among all implemented machine learning algorithms the Random Forest classifier trained with features selected using the Linear SVM algorithm achieved the highest classification accuracy of 73.4 

Reference \cite{Deschamps2021} aims at developing an automated pre-diagnosis tool for dysgraphia allowing a wide screening among children. Multiple tablets were utilized for data collection to ensure that the proposed system is not tablet/data collection tool dependent. To the best of our knowledge, this work has created the largest database for dysgraphia diagnosis by participating the 580 children for data collection.  The children from 2nd to 5th grade are considered for the experiment. And among them, 122 children are having dysgraphia.  The children were asked to perform BHK ( French version) on the tablet. 100 different features were extracted from the handwritten task.  80 \% of the feature samples are utilized to train the SVM machine learning classifier and the remaining 20\% has been used for testing. In the test set, the SVM classifier achieved a sensitivity of 91\% and specificity of 81\% for the detection of children with dysgraphia. 

The performance of multiple machine learning algorithms for the detection of dysgraphia is analyzed in \cite{Et.al.2021} . Initially, a dataset was constructed by conducting various handwritten-based exercises in a group of 240 subjects where 142 have some sort of learning disabilities and 45 subjects have dysgraphia. These subjects were asked to complete a dot connecting exercise to analyze their motor skill. Furthermore, the samples of writing are analyzed for legibility and space knowledge. Boolean features are extracted for the following properties: sentence structure, word formation, visual-spatial response. Along with that, the images of writing are subjected to SSIM evaluation, and spellings are checked through a spell checker. Among all the extracted features the relevant features are selected using a feature selection technique known as Elastic Net\cite{Jenul2021}. The selected relevant features are used to train KNN \cite{Zhang2016}, Naïve Bayes, Decision tree\cite{Quinlan1986} , Random Forest, SVM classification model for prediction of dysgraphia. The Random forest classifier achieved the highest classification accuracy and it can differentiate the dysgraphic and non-dysgraphic subjects with an accuracy of 99\% in the experimented dataset. 

A mobile-based system named ‘Nana Shilpa’ is proposed for screening and identifying the risk of dysgraphia as well as dyscalculia in primary school students \cite{Hewapathirana2021}. The proposed software solution is targeted at primary school students in Srilanka. For dysgraphia screening, the proposed method predicts the risk using two risk identification components letter level dysgraphia and numeric dysgraphia. To identify the risk of letter level dysgraphia the students were asked to write 4 letters and 4 words which are preselected by primary school teachers and the teacher’s guide of Srilanka. A convolutional neural network model which consists of four two-dimensional convolutional layers are developed for initial screening of letter level dysgraphia. The letter and word data images collected from students are used to train the CNN model. Before training, the image data are subjected to preprocessing to segment the letters and words. The CNN model will predict the letter as well as words along with the confidence score. To identify the risk, the output of the CNN model including the letter level accuracy confidence of written letters, words, total correct count of letters and words, along with the total number of erase counts are given as features to train an SVM model. The trained SVM model will predict the risk of letter level dysgraphia.  For numeric dysgraphia screening, the students were asked to write the number from 0 to 9 in the drawable canvas of the application. The correctness of the written number, erase count, time taken are considered as the features for the prediction of risk. The correctness or accuracy of the written number is determined using a CNN model trained in the MNIST dataset. These features are then used to train a machine learning model to predict the risk of numeric dysgraphia. The proposed methods can predict the risk of numeric and letter level dysgraphia with an accuracy of 99

A random forest-based machine learning method for dysgraphia identification from digitized handwriting is proposed in \cite{Dankovicova2019}. A dataset was constructed for handwritten analysis by making 78 subjects do various handwriting tasks in a Wacom Intous Pro Large graphic tablet. Among 78 subjects, 36 were having dysgraphia disability. Every handwriting was carefully analyzed by three specialists to create the ground truth labels. Various kinematic, temporal, and other features such as pen lifts, etc . are extracted for each written sample for further analysis. Three different machine learning classifiers Random Forest, SVC, and AdaBoost are trained using the extracted features for automated detection of dysgraphia. The obtained results indicate that Random forest displayed better classification performance. 

A software system based on the SVM method and Android application is proposed for dysgraphia identification from handwriting data in elementary school students \cite{Sihwi2019}. An Android application employed with a handwriting recognition tool namely “WritePad” is used for data acquisition. The students were asked to write on the screen of the smartphone and the following data: time, pressure, spacing between the letters, size of letters, position of letters, consistency of boundaries, etc. are stored for further analysis. The ground-truth labels of classes include three classes of dysgraphia light, moderate, and severe as well as normal are constructed using an existing method proposed in \cite{Kurniawan2018}.  SVM classifiers are implemented in two different ways: OVO and OVR approach for classification. The result, after using three different kernels in SVM such as Linear, Polynomial, and Radial Base Function kernel (RBF), shows that the RBF kernel produces better average accuracy and Cohen's kappa value compared to Linear and Polynomial kernels, where the average accuracy of each kernel is 78.56\% for Linear, 81.40\% for Polynomial, and 82.51\% for RBF.

Rosenblum et al. \cite{rosenblum2016identifying} proposed an online handwritten feature-based tool for the diagnosis of dysgraphia in 3rd-grade students. 99 students have participated in the study where 50 students are proficient in handwriting. The handwriting proficiency screening questionnaire (HPSQ) approach is utilized to classify the students into two groups ( with handwriting difficulties and without handwriting difficulties). The students are asked to do multiple tasks in a paper attached to a tablet to collect the online handwritten data.  ComPET tool is utilized for data collection and analysis.  Initially, the students were asked to write a six-word sentence from a well-known children's in the Hebrew language.  Different forms of time features, static properties of written product, and pressure features were extracted during this task for each student. In the second task, the students were asked to write a different six-word sentence. The features obtained in the second task are highly correlated with the features obtained in the second task. Thus the features obtained in the second task are omitted from the dataset. In the third task, students were asked to write a single Hebrew letter seven to ten times continuously. From that total time as well as the total number of segments in the written and total number of completed letters are extracted as the features.  The fourth task is associated with the doodling of connected loops. Time, as well as the geometry of the loops produced, are extracted from the data to build features. The extracted features are utilized to train an SVM linear model for classification.  The classification results indicate that the proposed methodology classifies the students with dysgraphia and without dysgraphia with a specificity and sensitivity of 90\%. Furthermore, the proposed study found that time-related features and pressure-related features are having high relevance in discriminating the normal and abnormal handwritings. 

Zvoncak et al. \cite{zvoncak2018effect}proposed a novel intra-writer normalization approach for reducing the error rate in computer-based development dysgraphia diagnosis. The data for the study was collected from 97 students by making them write a paragraph on a paper affixed to the screen of the digitizing tablet. The ground truth value is labeled using the HPSQ approach. Common spatial, kinematic, temporal, and dynamic features were extracted for each student during the handwriting task. Four intra-writer normalization approach, l1, l2, l infinity, and Z score has been proposed in this work. Instead of classification, a regression task is performed in this work to estimate the effectiveness of the normalization methods. A gradient boosted tree-based regression model is trained in this work for estimating the handwriting quality scores and it is validated against the ground truth value obtained during the HPSQ approach. Various experiments were conducted to analyze the effectiveness of different feature combinations and normalization methods. Observed results indicate that the l2 normalization can reduce the error rate by 5\%.  

A deep learning-based mobile application named “Pubudu” \cite{kariyawasam2019pubudu}is proposed for screening and intervention multiple learning disorders like dyslexia, dysgraphia, and dyscalculia in school children. In dysgraphia diagnosis, the proposed application is targeted for screening numerical as well as letter dysgraphia. In the letter dysgraphia screening approach, a CNN model is trained with the 5000 handwritten image data ( 3 Sinhala letters) collected from non-dysgraphic children. After that, the students with dysgraphia and without dysgraphia were asked to write the same letters, and multiple features such as success probability of written letters, the total number of correct letters, total number of incorrect letters, number of attempts, total time take as well as erase count were extracted from those handwritten data.  The trained CNN model is used for automated estimation of the success probability of the written letter, the total number of correct letters, and the total number of incorrect letters. These extracted features are used to train an SVM model for letter dysgraphia screening. A similar approach was employed for numerical dysgraphia screening, instead of the customized letter dataset, the MNIST dataset was used to train the CNN model for automated feature extraction purposes. The proposed approach obtained an accuracy of 88\% and 90\% for screening letter dysgraphia and numerical dysgraphia. 

A deep learning-based method is proposed in \cite{yogarajah2020deep} for automated detection of dyslexia-dysgraphia from handwritten images. For dataset preparation, an initial screening was conducted among the student of 1st grade to 5th grade to find out the students with poor scholastic records. From that 54 students were identified as having dyslexia-dysgraphia with the help of a conventional approach. Instead of making these students write new words or draw something, handwritten samples from their Hindi notebooks are collected, and images of some predefined two-letter words and vowel signs are extracted to form the dataset.   The dataset consists of a total of 267 handwritten images where 164 images are of class dyslexia-dysgraphia and 103 are of class normal. The normal images are collected from students of similar age without any learning disabilities. The images are converted to grayscale format and resized to a fixed height of 113 pixels. A convolutional neural network model with three 2D convolutional layers is trained with the random patches of the images for automated diagnosis. The developed CNN model achieved a classification accuracy of 86.14 \%. 

Devi et al. \cite{devi2021early}proposed a learning disability diagnosis tool for children with age 7 to 8 by utilizing the decision tree algorithm. The proposed tool can diagnose the existence of different learning disabilities including dysgraphia, dyscalculia, dyslexia, etc. An E-learning framework named testing scale is constructed for data collection and examination. The children have to answer a set of questionnaires to validate the existence of a learning disability. The questions are extracted from the WOODCOCK JOHNSON test of cognitive abilities. For dysgraphia, the set of questions is related to visual auditory, drawing lines, figure tracing, matching shape encircling, etc. The response of the students are extracted and features are created to train the decision tree algorithm. Only 40 students have participated in the experiment for data collection and evaluation. 

A digital handwriting analysis-based system for dysgraphia identification is proposed in \cite{kedar2021identifying}. Similar to other works, the proposed method makes use of a Wacom digital tablet affixed with white paper and an interactive stylus pen for data collection. The participating students were asked to perform five different tasks and the total time, tip pressure, X coordinate, and Y coordinate of the pen tip on the paper at each instant during the task are stored in the device. From the stored data nine different feature values such as the total number of on paper segments, in air segments, total time, total air time, total time spent on paper, mean pressure, the velocity of the pen, etc. were extracted for further analysis. Three different machine learning classifiers are trained using the extracted features. The first and second classifier is based on the decision tree and random forest algorithm respectively. The third classifier is an ensemble model based on a soft voting approach. The ensemble classifier consists of three separately trained individual classifiers: decision tree, random forest, and SVM. A total of 60 students have participated in the proposed work for dataset construction. In the constructed dataset the Random forest classier achieved a maximum  accuracy of 92.59\%. 

Comparative analysis of few relevant ML based dysgraphia screening systems are provided in table \ref{tab:4}

\begin{landscape}[]

\begin{longtable}{|p{0.6cm}|p{1.7cm}|p{1.7cm}|p{1.7cm}|p{1.7cm}|p{1.7cm}|p{1.7cm}|p{1.7cm}|p{4cm}|}

		\caption{Comparative analysis of machine learning based dysgraphia diagnosis systems}\label{tab:4}\\
		
		\hline
		Ref.     & Age group         & Participants                             & Tools                                                                          & Task                                                               & Features                                                                                                       & ML Algorithm                                             & Performance           & Remarks                                                                                                                                                                                                                                                                                                                                                                                                                                \\ \hline
		\cite{Mekyska2017} & 3rd grade         & 54 subjects ( 27 are having dysgraphia)  & WACOM Intuos II xy digitizing tablet, Model GP-110 wireless electronic pen     & Write a sequence of seven semi-HET (letters)                       & Kinematic measures, 34 nonlinear dynamic, other 7 Features                                                     & Linear discriminant analysis, Random forest              & Sensitivity : 96\%    & (+)Altitude/tilt and pressure features discriminate well                                                                                                                                                                                                                                      \\ \hline
		\cite{Asselborn2018} & Primary school    & 298 subjects (56 are having dysgraphia)  & Wacom graphic tablet                                                           & Copying a text for 5 mins                                          & Static,  Kinematic, Pressure, Tilt                                                                             & Random forest                                            & Sensitivity : 96.5 \% & -                                                                                                                                                                                                                                                                                                                                                                                                                                      \\ \hline
		\cite{Isa2019}  & 7 - 12 years      & -                                        & Scanner                                                                        & Samples of handwritten text                                        & OCR,MSER                                                                                                       & Artificial neural network                                & Accuracy : 71 \%      & (-) Only image of handwritten text , which is suitable for spatial dysgraphia screening alone.                                                                                                                                                                                                                                                                                                                                         \\ \hline
		\cite{Mekyska2019} & 3rd and 4th grade & 76 subjects ( 15 are having dysgraphia)  & Wacom Intuos Pro L (PHT-80) tablet                                             & Seven drawing activities                                           & Spatial, temporal, kinematic, dynamic, other – pen elevations and relative number of interruptions             & XG-Boost                                                 & Specificity : 90\%    & (+)  Kinematic features are found to be more relevant for classification , specifically the kinematic features in vertical projection.                                                                                                                                                                                                                                                                                              \\ \hline
		\cite{Zvoncak2019} & 3rd and 4th grade & 65 subjects ( 33 are having dysgraphia)  & Digitizing tablet Wacom Intuos Pro L (PHT-80) digitizer with Wacom Inking pen. & Copy a short paragraph (63 words, 371 characters including spaces) & Kinematic, temporal, spatial, and dynamic                                                                      & Support vector machine and Random forest                 & Sensitivity : 88\%    & (+) The features from vertical movement and pressure are having highest discrimination power. It indicates that subject with dysgraphia are having difficulties in maintaining constant force on the pen tip and their vertical movement are less fluent  \\ \hline
		\cite{dui2020tablet} & Pre-schoolers     & 104 subjects ( 28 are having dysgraphia) & An application in Unity 2018.3.2f1, for an iPad 6, with Apple Pencil           & Copy square, copy sequence of symbols, Questionare                 & Gesture smoothness, pressure(mean value), Drawing kinematics                                                   & Logistic regression                                      & AUC : 0.82            & (-) Since done in preschool stage, it is not possible to confirm that real handwriting difficulties would arise in the following year school stage                                                                                                                                                                                                                                                                                     \\ \hline
		\cite{Gargot2020}  & Around 9 years    & 280 subjects (62 are having dysgraphia)  & Sheet of paper affixed to a Wacom Intuos 4 graphic tablet                      & -                                                                  & Static, kinematic, pressure, and tilt                                                                          & Multiple linear regression models and K-means clustering & -                     & (+) Pressure on the pen tip and tilt features were found to be significant                                                                                                                                                                                                                                                                                                                                                             \\ \hline
		\cite{Drotar2020} & 8–15 years        & 120 subjects                             & WACOM Intuos Pro Large tablet                                                  & Writing letters, words, sentence                                   & Spatiotemporal and kinematic features                                                                          & AdaBoost                                                 & Accuracy : 80\%       & (+)Pressure of pen tip on the surface and pen lifts were among the features with high discriminatory potential . The dataset includes subjects aged 8–15 years, which is basically a broad range and it has affected the overall accuracy.                                                                                                                                                  \\ \hline
		
		\hline
		\cite{rosenblum2016identifying} & 8–9 years        & 90 subjects (49 are non-proficient ) subjects                             & WACOM Intuos II x-y digitizing tablet     & Writing letters, words, sentence                                   & Spatiotemporal , dynamic, kinematic and other features                                                                          & Support vector machines                                                & Accuracy : 90\%       & (+)Pressure of pen tip on the surface and time were among the features with high discriminatory potential .                                                                                                                                                \\ \hline

		\cite{Sihwi2019} & 3rd to 6th grade        & 32 subjects               &Smartphone  & Drawing and writing                           & Spatial, temporal , dynamic and other features                                                                          & Support Vector machines                                               & Accuracy : 82.51\%       & (-)   Less subjects
		(+) Instead of binary classification , the proposed system can classify the data to different levels of dysgraphia such as mild , moderate , severe etc.   \\ \hline
		
		\cite{Dankovicova2019}& 10-13 years        & 72 subjects (36 are having dysgraphia )                            & WACOM Intuos Pro large     & Writing letters, words, sentence                                   & Spatial,temporal , dynamic, kinematic and other features                                                                          & Support Vector machines                                               & Sensitivity : 75.5\%       & (-)   Less subjects
		(-) Not an acceptable sensitivity \\ \hline
		
		\cite{devi2021early} & 7-8 years       & 40 subjects                            & A web based framework                                                  & drawing lines, tracing figures    , encircling matching shapes                               & Not explicitly mentioned                                                                       & Decision tree                                              & Not mentioned      & (-) Number of data samples  are very less. (-) Children with other learning disorders such as dyscalculia , dyslexia are also included in the experiment\\ \hline
		
		\cite{kedar2021identifying} & 7-12 years        & 60 subjects                             & WACOM digital tablet                                                  & Writing and Drawing  .                                 & Spatiotemporal, dynamic and kinematic features                                                                          & Random forest                                                & Accuracy : Sensitivity 92.85\%       & (-) Number of data samples ( only 60 ) are less compared to other works.                                                                                                                                                  \\ \hline
		
		\cite{Asselborn2020} & 5 - 12 years      & 448 subjects (58 are having dysgraphia)  & iPad tablet                                                                    & Write the five first sentences used in the BHK test                & Pressure, tilt and azimuth angle of the pen                                                                    & PCA and unsupervised methods                             &                       & (+) Kinematic and pressure features are found to be more important than static or tilt features. (+)Age treated as  a continuous variable since it is evident that handwriting evolves at a more rapid pace. (+)Handwriting quality measured on a numeric scale, allowing a new categorization of handwriting difficulties                                                          \\ \hline
		\cite{Deschamps2021} & 2nd to 5th grade  &           580 subjects (128 are having dysgraphia)                               & Four different graphic tablets                                                                 & Perform french version of BHK test on tablet.                                                & 100 features including spatial ,kinematic , dynamic etc. &    SVM                                                      &                Sensitivity :91\%       & (+) Different types of tablets are used for data collection to ensure that the data is not device dependent .                                                                                                                                                                                                                                                                                                                               \\ \hline

\end{longtable}

\end{landscape}

\subsection{Automated Diagnosis Systems Without ML}

Non machine learning  based automatede digital screening systems are also proposed in the literature for diagnosing the dysgraphia. This section briefly discus about popular Non-ML based digital dysgraphia screening systems. A tablet-based app named Play Draw Write has been proposed in \cite{dui2020tablet} for screening the handwriting skills of children in the preliteracy stage. The proposed tablet app will quantify three handwriting laws isochrony, homothety, and speed-accuracy trade-off for assessing the existence of dysgraphia markers. “The isochrony principle states that bigger gesture execution is accompanied by an increase in average movement speed to keep the movement time approximately constant \cite{viviani1982trajectory}. The homothety principle predicts that the fraction of time devoted to each letter of a word is kept constant and is independent of the total word duration \cite{viviani1982trajectory,lashley1951problem}.” These principles seem to be altered in the subjects with dysgraphia along with dyslexia. The app is developed for Samsung Galaxy Tab A and children can use S-Pen with a rubber tip write or draw on the screen of the tablet App. Two tasks are available in the app, where the first one is a copy game and the second is a tunnel game. In the copy game, the students are asked to copy a few symbols and words to the canvas provided in the app using the stylus. The copying game is targeted for testing the isochrony and homothety. The tunnel game is targeted for evaluating the speed-accuracy trade-off. 

Giordano and Maiorana \cite{giordano2014addressing}have developed a web-based software system for multiple handwritten gesture recognition which can be extended for online dysgraphia screening. The proposed software system is designed in a client-server manner and it can be used in any type of smartphone or tablet and computers. A modified version of the dynamic wrapping algorithm is employed in the server to recognize multiple hand gestures. The proposed software system offers multiple functions to execute different writing and drawing tasks. For each task, the system will store the time taken, degree of similarity with the reference line, amount of points outside the reference line in terms of percentage, average horizontal distance, etc. From the stored data, the administrator of the software tool can derive other parameters such as time taken for each stroke or their average time, length of the path, velocity, total air time, trajectory analysis, etc. These features can be utilized for the development of machine learning classifiers for dysgraphia diagnosis. 

Raza et al.\cite{raza2017interactive} proposed an interactive mobile application for dysgraphia screening in children from age 5 to 12. The mobile application offers different activities to assess the handwriting quality of the student, assess the phonological dysgraphia, assess the surface dysgraphia, assess the copying ability. The student can use any compatible stylus pen to write on the screen of the device. The software utilizes a handwritten recognition method to recognize the spelling and after recognition, it is compared with the ground truth value to provide a score based on the correctness. A total of twenty words are given for students to write where seven words are used to estimate the existence of phonological dysgraphia, eight words for surface dysgraphia, and the remaining for copying ability. 

TestGraphia \cite{Dimauro2020}is a software system proposed for the early diagnosis of dysgraphia. The proposed system is based on the conventional BHK method. They extended the conventional BHK method to the software device for automated diagnosis. In the BHK test, the students are asked to copy a few given texts to a paper in 5 minutes. From the handwritten document, scores are given to the 13 features including writing size, skewed writing, sharp angles, etc., and scoring these features is the crucial step in the BHK test. The TestGraphia will automatically compute the scores for 9 features  ( "non-aligned left margin", "skewed writing" , "insufficient space between two words" , "sharp angles" , " broken links between two letters" , "irregular size of letters", " inconsistent height between letters with extension and letters without extension",  "ambiguous letters"). Scores of the remaining features can be added by the doctors or occupational therapists manually using the dashboard of the software. For the automated scoring, initially, the images of the writing are subjected to segmentation ( individual line segmentation) using an image processing algorithm based on horizontal and vertical histograms. Multiple simple image processing algorithms are employed on the segmented lines to compute the feature score. The proposed system was tested with 109 students ( 2nd – 5th grade). The obtained result indicates that TestGraphia can diagnose the existence of dysgraphia with a sensitivity value of 0.83. 

\subsection{Commercial Systems for dysgraphia diagnosis}

 In this section, we briefly discuss the available software systems in the marketplace for dysgraphia diagnosis. There are very few commercially available systems for dysgraphia screening. ‘Lexcercise’ is a company founded in 2008 for supporting students with learning disabilities. Their web application provides an easy and cost-free way for preliminary assessment of dysgraphia in children \cite{Le}. The whole procedure for assessment consists of three sections. The parents or caretakers of the student are asked to fill out the answers to the provided questions about the student. In the first section, the questions are related to letter and number writing (such as reverse writing, messy writing, mixing up of lowercase and uppercase). The questions in the second section and third sections are targeted for assessing the writing convention and writing proficiency of the student respectively. They will provide the service of the dysgraphia therapist also. Upon completion of the assessment, the parents or caretaker will be asked to submit the actual writing samples of the student to a dysgraphia therapist, if the assessment displays any risk.  \cite{Addit} also, provide a web-based tool for the free assessment of dysgraphia. It is similar to \cite{Le}, just a screener test to decide whether the child is having the symptoms of dysgraphia. The assessment consists of 16 basic questions where most of which are considered to be the prevalent symptoms of dysgraphia.  All the questions are prepared based on the criteria from the learning disabilities association of America. Unlike \cite{Le}, here they will not provide the therapist, but they will give suggestions for the parents to check with an occupational therapist if the test is positive. 

Dyscreen \cite{Dy} is a smartphone application developed by the Australian company Dystech in 2020 for self-screening of dysgraphia and dyslexia using the power of artificial intelligence. Dystech claims that they are the pioneer in this field to introduce the first AI-based smartphone application for dysgraphia screening. Currently, the application is available for both iOS and Android. The dysgraphia screening functionality of the application is very simple. Just take the picture of the student's handwritten text and upload it via the application. The intelligent machine learning model deployed in the cloud will classify the handwritten sample into positive or negative classes and provide feedback to the user.  Since the processing of the images is done in the cloud, internet connectivity is required for dysgraphia screening. Dystech claims that the machine learning model can predict the existence of dysgraphia with an accuracy of 95.6\%. The dysgraphia screening in Dyscreen is free of cost. The same application provides other services such as dyslexia screening, but it is not free of cost. 
A team of professionals and engineers from Hongkong university of science and technology has developed an artificial intelligence-based smartphone application named “ AI Dysgraphia Pre-screening’ for dysgraphia screening \cite{AIdy}.  The application is free and available on both Android as well as iOS. The application provides a 15-minute drawing exercise for the student. Upon completing the drawing exercise the essential dynamic and static features are extracted from the handwritten data and processed for data analysis using trained ML algorithms. Internet connectivity is required to access the app services. Compared to Dyscreen this app considered the dynamics of writings also for screening dysgraphia. 

To the best of our knowledge, these are the available software tools for dysgraphia screening in the market. All are free to access, however have many limitations. The Dyscreen just considers the images only, so the dynamics of the handwriting are not considered. On the other hand, in \cite{AIdy} the data are collected via smartphone screen, and it is very different from a normal paper, and friction while writing will be very different.  The marketplace lacks a sophisticated smart device that can intelligently detect dysgraphia.

\section{Discussion and Future recommendations}

The popularity and the built-in capabilities of digitizing tablets to acquire information such as the position of the pen tip, on surface/  in-air pen position, pen tip pressure, the azimuth angle of pen with respect to the tablet surface, the tilt of pen, timestamp \cite{Mekyska2017} made it a suitable tool for the handwriting analysis.  Stroke dimensions, Velocity, Acceleration, Jerk, Pressure, Tilt, Temporal, Azimuth angle, Number of pen elevations are the features that have more or less equal prominence in the dysgraphia discrimination systems \cite{Mekyska2017,Asselborn2018,Mekyska2019,Zvoncak2019,Dui2020,Drotar2020,Asselborn2020,Gargot2020} . The most relevant features that are spotted out in a multitude of works are Kinematic and Pen tip pressure. 
Compared to the approaches merely based on handwriting images,\cite{wu2019automated,hen2019characteristics} the tablet-based techniques could explore more characteristics of handwriting, which turned out to be significant for the detection of dysgraphia. The data acquisition in the latter approach involved writing with a normal pen or an electronic pen on paper overlaid on the tablet. However, the lower friction surface of tablet computers modifies graphomotor execution, which in turn contradicts the purpose \cite{Guilbert2019}. Moreover, the pressure sensitivity of these tablets may vary depending on the model \cite{Prunty2020}. Even though the grip force between the hands and writing instrument has a substantial correlation to the fine motor performance \cite{Lin2017} and the improper handling of the pen is prevalent in motor dysgraphia \cite{Biotteau2019}, it is not considered as an attributing factor for dysgraphia prediction. The existing methodologies are meant to provide an overall diagnostic assessment as being dysgraphic/ non-dysgraphic except a few with a grading scale for handwriting difficulties . In addition, a single task (copying/ writing/ drawing) is not sufficient for the judgment .

In our knowledge there are number of gaps in the automated dysgraphia detection system domain. Some of the recommendations for future works are : 

\begin{itemize}
\item Since the pressure sensitivity of the tablets varies by models and tablets have lower friction surfaces, future dysgraphia diagnosis systems can consider normal paper (writing on paper) based systems rather than writing on tablets.

\item The future systems can explore the importance of improper handling of pen/pencil and grip force between hand and writing instruments as relevant features for discriminating normal and abnormal handwriting. 

\item Instead of developing binary systems ( dysgraphia or not)  for dysgraphia diagnosis, future work should consider developing systems for recognizing different types of dysgraphia or grading the levels of dysgraphia. Because the symptoms of dysgraphia vary with different types and levels of disability.

\end{itemize}

\section{New frameworks  for dysgraphia daignosis}

\begin{landscape}

\begin{figure}[h]%
\centering
\includegraphics[width=1.6\textwidth]{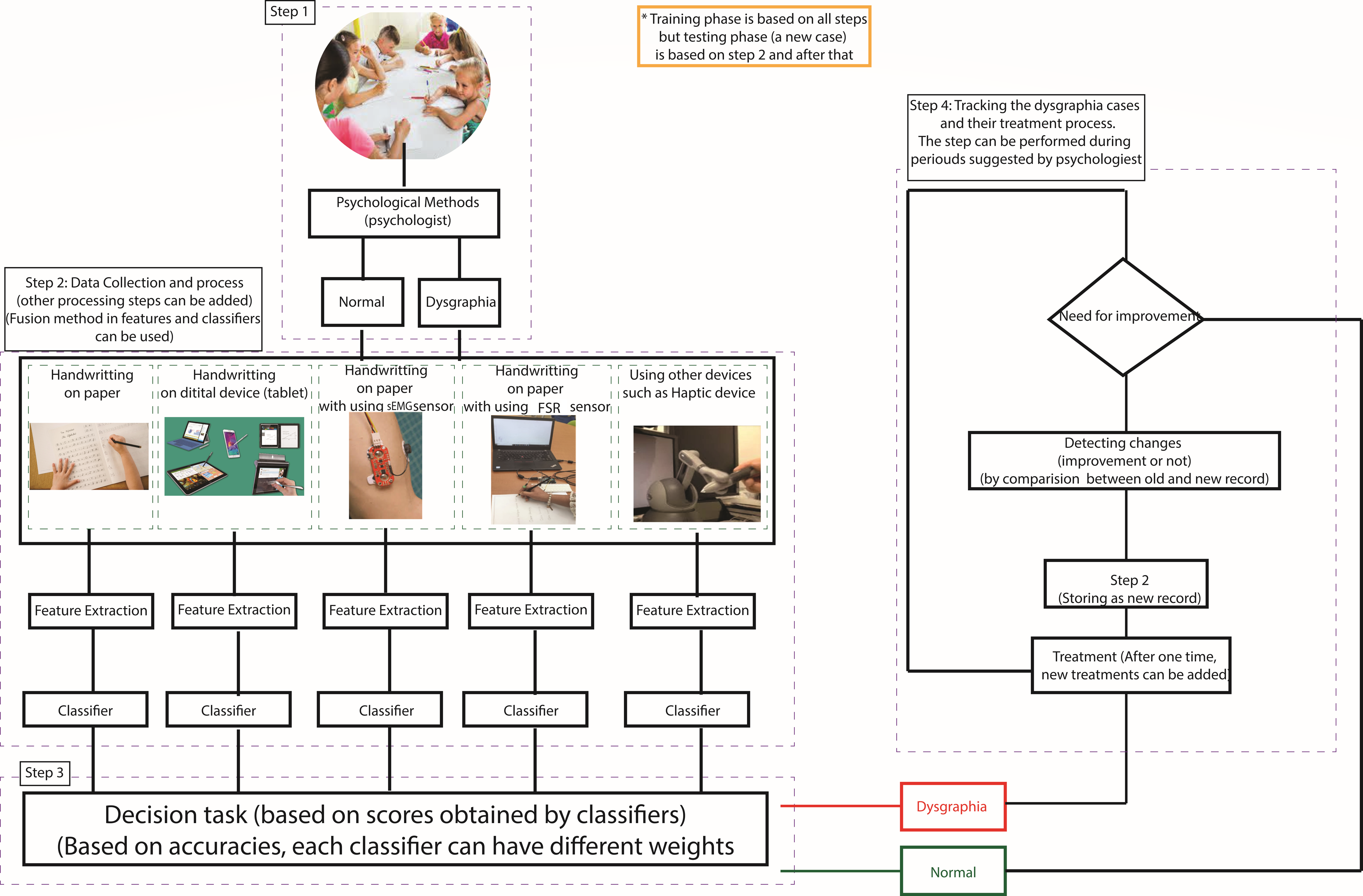}
\caption{Overview of proposed framework}\label{fig7-1}
\end{figure}

\end{landscape}

To address some of the limitations in the existing frameworks, we propose new frameworks for dysgraphia diagnosis. Overview of the framework is shown in figure \ref{fig7-1}. The framework consists of 4 steps. In the first step, children are classified into normal and dysgraphic children using psychological methods. Then, an automatic system is developed that takes into account data collection and processing. In this step, depending on the type of approach used, other processing steps can be added. For example, for handwriting on paper, preprocessing should be added, and a fusion method for features and classifiers can be used. Furthermore, instead of
 feature extraction and classifier, deep learning methods can also be used. In the collection step, the content that the participants are asked to write or draw is very important. Depending on what equipment is available, a combination of this equipment can also be considered. In feature extraction, off-line and on-line features are considered based on the approach of data collection.

It should be noted that step 1 is considered to obtain training data. In step 3, you can start adding test data to evaluate the system based on the trained model and decide to which classes the new case (test data) belongs. In this step, we can assign different weights to each classifier depending on the type of data collection. For example, since handwriting on paper can lead to low accuracy, we consider a lower weight. In the last step, which is very important, an algorithm is proposed for tracking the dysgraphia cases 
 and their treatment process.
This step can be carried out in the periods proposed by the psychologist. 
After dysgraphia cases are detected, treatment should be performed. Impact of the treatment method can be explored during the tracking step. It is possible to add new treatments and study their effectiveness.

\begin{figure}[h]%
\centering
\includegraphics[width=0.9\textwidth]{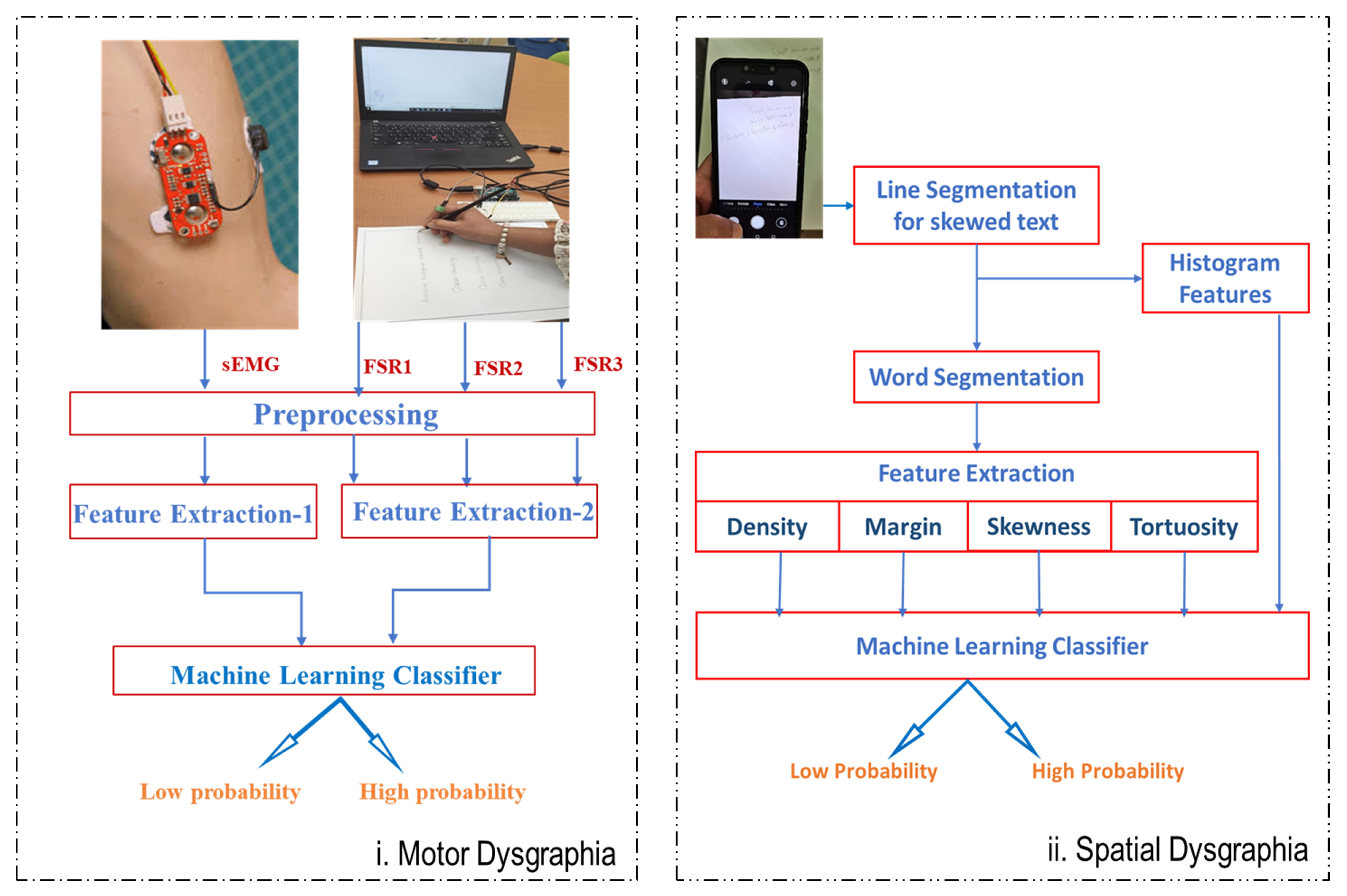}
\caption{Work flow of novel methods}\label{fig7}
\end{figure}

It should be noted that some tasks and steps can be added to the framework such as detecting type of dysgraphia. The new framework consist of two  novel methods (figure \ref{fig7} ) that can be used along with existing methods for automatic screening of spatial dysgraphia and motor dysgraphia, respectively.For the former one, the dynamic characteristics of handwriting are captured with the help of two types of sensors- (1) Force-sensitive resistors (FSR) attached to the pencil for capturing pencil grip patterns (2) sEMG sensor attached to the hand for capturing muscle activity. The acquired data is preprocessed followed by feature extraction and binary classification with the help of prebuilt machine learning models. 

Handwriting is a prominent fine motor skill acquired during the early developmental stage. Several types of research have been carried out to understand the detailed aspects and characteristics of handwriting. Earlier in 1961, Herrick and Otto have signified the barrel pressure variability as an important distinguishing aspect of individual handwriting \cite{herrick1961pressure}. The grip force variability and its effect on handwriting legibility are studied in \cite{falk2010grip}. The demonstrated results indicated that grip force variability during the entire writing task was lower for non-proficient writers. Reference \cite{lin2017comprehension} investigated the pen grip kinetics of school-age children to validate the hypothesis of correlation between force control when handling a pen and fine motor performance. The significance of the middle finger compared to the thumb and index finger for the force control is also demonstrated. The writing speed being unaffected by the pencil grip force and muscular activity holds promise for the application of these cues in the remedial program for disabled persons.Surface Electromyography (sEMG) has proved out to study muscle coordination \cite{hug2018surface}, which can be exploited to understand the poor muscle tone in motor dysgraphia. sEMG signals generated during handwriting have been already applied for character recognition \cite{linderman2009recognition,beltran2020multi}and Parkinson’s disease diagnosis \cite{loconsole2019model}. It has been used in Arabic handwriting character recognition as well \cite{lansari2003novel}. 

For the latter one, a mobile phone camera is utilized to capture the image of handwritten piece. Relevant features are extracted from the image and further classification is carried out with the aid of machine learning models. The illegibility aspect of handwriting due to the defect in understanding of space results in spatial dysgraphia. Suitable line/ word segmentation techniques \cite{alma2006recognition} can be utilized to observe and compare the line as well as the word spacing.  Further, document analysis algorithms \cite{Mekyska2017} serves to identify the dysgraphic characteristics such as skewed writing, irregular writing size, etc.

\section{Conclusion}
Learning disabilities are often unrecognized, which makes it often misinterpreted as a lack of intelligence. This comes from the fact that the screening process is quite complex. In this work, we presented survey of existing  tools and method  for the preliminary screening of motor and spatial  dysgraphia, which is characterized by impaired handwriting skills. We discussed many AI based and Non AI based automated systems for dysgraphia screening. We provided a comparative analysis of these systems and highlighted the strength and weakness . Later we proposed a novel frame work for automated dysgraphia screening by combining the existing methods along with new methods .

\section*{Declarations}

\begin{itemize}
\item Funding
 This publication was supported by Qatar University Graduate Assistant
Grant. The contents of this publication are solely the responsibility of the
authors and do not necessarily represent the official views of Qatar University.

\item Conflict of interest/Competing interests 
The authors declare that they have no competing interests.

\item Ethics approval 
Not applicable

\item Consent to participate
Not applicable

\item Data availability
Data sharing not applicable to this article as no datasets were generated or analysed during the current study.

\item Authors' contributions
JK have made major contribution in designing the work, reviewing the
literature and writing the manuscript. The concept of the work was contributed by SA and JK . SA was the principal investigator of the project.
SA as well as YA revised and edited the manuscript.SK provided guidance
and as well helped in drafting the manuscript. All authors read and approved
the final version of manuscript.

\end{itemize}


\bibliography{MyCollection}


\begin{thebibliography}{101}
\ifx \bisbn   \undefined \def \bisbn  #1{ISBN #1}\fi
\ifx \binits  \undefined \def \binits#1{#1}\fi
\ifx \bauthor  \undefined \def \bauthor#1{#1}\fi
\ifx \batitle  \undefined \def \batitle#1{#1}\fi
\ifx \bjtitle  \undefined \def \bjtitle#1{#1}\fi
\ifx \bvolume  \undefined \def \bvolume#1{\textbf{#1}}\fi
\ifx \byear  \undefined \def \byear#1{#1}\fi
\ifx \bissue  \undefined \def \bissue#1{#1}\fi
\ifx \bfpage  \undefined \def \bfpage#1{#1}\fi
\ifx \blpage  \undefined \def \blpage #1{#1}\fi
\ifx \burl  \undefined \def \burl#1{\textsf{#1}}\fi
\ifx \doiurl  \undefined \def \doiurl#1{\url{https://doi.org/#1}}\fi
\ifx \betal  \undefined \def \betal{\textit{et al.}}\fi
\ifx \binstitute  \undefined \def \binstitute#1{#1}\fi
\ifx \binstitutionaled  \undefined \def \binstitutionaled#1{#1}\fi
\ifx \bctitle  \undefined \def \bctitle#1{#1}\fi
\ifx \beditor  \undefined \def \beditor#1{#1}\fi
\ifx \bpublisher  \undefined \def \bpublisher#1{#1}\fi
\ifx \bbtitle  \undefined \def \bbtitle#1{#1}\fi
\ifx \bedition  \undefined \def \bedition#1{#1}\fi
\ifx \bseriesno  \undefined \def \bseriesno#1{#1}\fi
\ifx \blocation  \undefined \def \blocation#1{#1}\fi
\ifx \bsertitle  \undefined \def \bsertitle#1{#1}\fi
\ifx \bsnm \undefined \def \bsnm#1{#1}\fi
\ifx \bsuffix \undefined \def \bsuffix#1{#1}\fi
\ifx \bparticle \undefined \def \bparticle#1{#1}\fi
\ifx \barticle \undefined \def \barticle#1{#1}\fi
\bibcommenthead
\ifx \bconfdate \undefined \def \bconfdate #1{#1}\fi
\ifx \botherref \undefined \def \botherref #1{#1}\fi
\ifx \url \undefined \def \url#1{\textsf{#1}}\fi
\ifx \bchapter \undefined \def \bchapter#1{#1}\fi
\ifx \bbook \undefined \def \bbook#1{#1}\fi
\ifx \bcomment \undefined \def \bcomment#1{#1}\fi
\ifx \oauthor \undefined \def \oauthor#1{#1}\fi
\ifx \citeauthoryear \undefined \def \citeauthoryear#1{#1}\fi
\ifx \endbibitem  \undefined \def \endbibitem {}\fi
\ifx \bconflocation  \undefined \def \bconflocation#1{#1}\fi
\ifx \arxivurl  \undefined \def \arxivurl#1{\textsf{#1}}\fi
\csname PreBibitemsHook\endcsname

\bibitem{Knickenberg2020}
\begin{barticle}
\bauthor{\bsnm{Knickenberg}, \binits{M.}},
\bauthor{\bsnm{{L. A. Zurbriggen}}, \binits{C.}},
\bauthor{\bsnm{Venetz}, \binits{M.}},
\bauthor{\bsnm{Schwab}, \binits{S.}},
\bauthor{\bsnm{Gebhardt}, \binits{M.}}:
\batitle{{Assessing dimensions of inclusion from students'
  perspective–measurement invariance across students with learning
  disabilities in different educational settings}}.
\bjtitle{European Journal of Special Needs Education}
(\byear{2020}).
\doiurl{10.1080/08856257.2019.1646958}
\end{barticle}
\endbibitem

\bibitem{Lyon1995}
\begin{botherref}
\oauthor{\bsnm{Lyon}, \binits{G.R.}}:
{Toward a definition of dyslexia}.
Annals of Dyslexia
\textbf{45}(1)
(1995).
\doiurl{10.1007/BF02648210}
\end{botherref}
\endbibitem

\bibitem{Keong2020}
\begin{botherref}
\oauthor{\bsnm{Keong}, \binits{W.K.}},
\oauthor{\bsnm{Pang}, \binits{V.}},
\oauthor{\bsnm{Eng}, \binits{C.K.}},
\oauthor{\bsnm{Keong}, \binits{T.C.}}:
{A framework for diagnosing dyscalculia}.
ASM Science Journal
\textbf{13}
(2020).
\doiurl{10.32802/ASMSCJ.2020.SM26(2.1)}
\end{botherref}
\endbibitem

\bibitem{Gibbs2007}
\begin{botherref}
\oauthor{\bsnm{Gibbs}, \binits{J.}},
\oauthor{\bsnm{Appleton}, \binits{J.}},
\oauthor{\bsnm{Appleton}, \binits{R.}}:
{Dyspraxia or developmental coordination disorder? Unravelling the enigma}
(2007).
\doiurl{10.1136/adc.2005.088054}
\end{botherref}
\endbibitem

\bibitem{Koravand2017}
\begin{botherref}
\oauthor{\bsnm{Koravand}, \binits{A.}},
\oauthor{\bsnm{Jutras}, \binits{B.}},
\oauthor{\bsnm{Lassonde}, \binits{M.}}:
{Abnormalities in cortical auditory responses in children with central auditory
  processing disorder}.
Neuroscience
\textbf{346}
(2017).
\doiurl{10.1016/j.neuroscience.2017.01.011}
\end{botherref}
\endbibitem

\bibitem{Deuel1995}
\begin{barticle}
\bauthor{\bsnm{Deuel}, \binits{R.K.}}:
\batitle{{Developmental Dysgraphia and Motor Skills Disorders}}.
\bjtitle{Journal of Child Neurology}
(\byear{1995}).
\doiurl{10.1177/08830738950100S103}
\end{barticle}
\endbibitem

\bibitem{Chung2020}
\begin{barticle}
\bauthor{\bsnm{Chung}, \binits{P.J.}},
\bauthor{\bsnm{Patel}, \binits{D.R.}},
\bauthor{\bsnm{Nizami}, \binits{I.}}:
\batitle{{Disorder of written expression and dysgraphia: definition, diagnosis,
  and management}}.
\bjtitle{Translational Pediatrics}
\bvolume{9}(\bissue{S1}),
\bfpage{46}--\blpage{54}
(\byear{2020}).
\doiurl{10.21037/tp.2019.11.01}
\end{barticle}
\endbibitem

\bibitem{LHamstra-BletzdeBieJ1987}
\begin{botherref}
\oauthor{\bsnm{{L, Hamstra-Bletz, de Bie J}}, \binits{d.B.B.}}:
{Concise evaluation scale for children's handwriting}.
Swets 1 zeitlinger ed.Lisse
(1987)
\end{botherref}
\endbibitem

\bibitem{Barnett2009}
\begin{barticle}
\bauthor{\bsnm{Barnett}, \binits{A.L.}},
\bauthor{\bsnm{Henderson}, \binits{S.E.}},
\bauthor{\bsnm{Scheib}, \binits{B.}},
\bauthor{\bsnm{Schulz}, \binits{J.}}:
\batitle{{Development and standardization of a new handwriting speed test: The
  Detailed Assessment of Speed of Handwriting}}.
\bjtitle{British Journal of Educational Psychology}
(\byear{2009}).
\doiurl{10.1348/000709909x421937}
\end{barticle}
\endbibitem

\bibitem{AmericanPsychiatricAssociation.2013}
\begin{bbook}
\bauthor{\bsnm{{American Psychiatric Association.}}},
\bauthor{\bsnm{{American Psychiatric Association. DSM-5 Task Force.}}}:
\bbtitle{{Diagnostic and Statistical Manual of Mental Disorders : DSM-5.}},
p. \bfpage{947}.
\bpublisher{American Psychiatric Association}, \blocation{???}
(\byear{2013})
\end{bbook}
\endbibitem

\bibitem{Lopez2018}
\begin{barticle}
\bauthor{\bsnm{Lopez}, \binits{C.}},
\bauthor{\bsnm{Hemimou}, \binits{C.}},
\bauthor{\bsnm{Golse}, \binits{B.}},
\bauthor{\bsnm{Vaivre-Douret}, \binits{L.}}:
\batitle{{Developmental dysgraphia is often associated with minor neurological
  dysfunction in children with developmental coordination disorder (DCD)}}.
\bjtitle{Neurophysiologie Clinique}
(\byear{2018}).
\doiurl{10.1016/j.neucli.2018.01.002}
\end{barticle}
\endbibitem

\bibitem{Gargot2020}
\begin{barticle}
\bauthor{\bsnm{Gargot}, \binits{T.}},
\bauthor{\bsnm{Asselborn}, \binits{T.}},
\bauthor{\bsnm{Pellerin}, \binits{H.}},
\bauthor{\bsnm{Zammouri}, \binits{I.}},
\bauthor{\bsnm{Anzalone}, \binits{S.M.}},
\bauthor{\bsnm{Casteran}, \binits{L.}},
\bauthor{\bsnm{Johal}, \binits{W.}},
\bauthor{\bsnm{Dillenbourg}, \binits{P.}},
\bauthor{\bsnm{Cohen}, \binits{D.}},
\bauthor{\bsnm{Jolly}, \binits{C.}}:
\batitle{{Acquisition of handwriting in children with and without dysgraphia: A
  computational approach}}.
\bjtitle{PLoS ONE}
\bvolume{15}(\bissue{9 September}),
\bfpage{1}--\blpage{22}
(\byear{2020}).
\doiurl{10.1371/journal.pone.0237575}
\end{barticle}
\endbibitem

\bibitem{rosenblum2018inter}
\begin{barticle}
\bauthor{\bsnm{Rosenblum}, \binits{S.}}:
\batitle{Inter-relationships between objective handwriting features and
  executive control among children with developmental dysgraphia}.
\bjtitle{PLoS One}
\bvolume{13}(\bissue{4}),
\bfpage{0196098}
(\byear{2018})
\end{barticle}
\endbibitem

\bibitem{vaivre2021phenotyping}
\begin{barticle}
\bauthor{\bsnm{Vaivre-Douret}, \binits{L.}},
\bauthor{\bsnm{Lopez}, \binits{C.}},
\bauthor{\bsnm{Dutruel}, \binits{A.}},
\bauthor{\bsnm{Vaivre}, \binits{S.}}:
\batitle{Phenotyping features in the genesis of pre-scriptural gestures in
  children to assess handwriting developmental levels}.
\bjtitle{Scientific reports}
\bvolume{11}(\bissue{1}),
\bfpage{1}--\blpage{13}
(\byear{2021})
\end{barticle}
\endbibitem

\bibitem{chang2013handwriting}
\begin{barticle}
\bauthor{\bsnm{Chang}, \binits{S.-H.}},
\bauthor{\bsnm{Yu}, \binits{N.-Y.}}:
\batitle{Handwriting movement analyses comparing first and second graders with
  normal or dysgraphic characteristics}.
\bjtitle{Research in developmental disabilities}
\bvolume{34}(\bissue{9}),
\bfpage{2433}--\blpage{2441}
(\byear{2013})
\end{barticle}
\endbibitem

\bibitem{vanithadyslexia}
\begin{botherref}
\oauthor{\bsnm{Vanitha}, \binits{G.}},
\oauthor{\bsnm{Kasthuri}, \binits{M.}}:
Dyslexia prediction using machine learning algorithms--a review
\end{botherref}
\endbibitem

\bibitem{chakraborty2019survey}
\begin{botherref}
\oauthor{\bsnm{Chakraborty}, \binits{M.V.}}:
A survey paper on learning disability prediction using machine learning.
INTERNATIONAL JOURNAL OF INFORMATION AND COMPUTING SCIENCE
\textbf{6}(5)
(2019)
\end{botherref}
\endbibitem

\bibitem{vanjari2019review}
\begin{bchapter}
\bauthor{\bsnm{Vanjari}, \binits{N.}},
\bauthor{\bsnm{Patil}, \binits{P.}},
\bauthor{\bsnm{Sharma}, \binits{S.}},
\bauthor{\bsnm{Gandhi}, \binits{M.}}:
\bctitle{A review on learning disabilities and technologies determining the
  severity of learning disabilities}.
In: \bbtitle{2nd International Conference on Advances in Science \& Technology
  (ICAST)}
(\byear{2019})
\end{bchapter}
\endbibitem

\bibitem{jothi2019prediction}
\begin{bchapter}
\bauthor{\bsnm{Jothi~Prabha}, \binits{A.}},
\bauthor{\bsnm{Bhargavi}, \binits{R.}}:
\bctitle{Prediction of dyslexia using machine learning—a research
  travelogue}.
In: \bbtitle{Proceedings of the Third International Conference on
  Microelectronics, Computing and Communication Systems},
pp. \bfpage{23}--\blpage{34}
(\byear{2019}).
\bcomment{Springer}
\end{bchapter}
\endbibitem

\bibitem{saxena2020machine}
\begin{botherref}
\oauthor{\bsnm{Saxena}, \binits{L.K.}},
\oauthor{\bsnm{Saxena}, \binits{M.}}:
Machine learning in diagnosis of children with disorders.
Machine Learning for Healthcare: Handling and Managing Data,
175
(2020)
\end{botherref}
\endbibitem

\bibitem{jamhar2019prediction}
\begin{bchapter}
\bauthor{\bsnm{Jamhar}, \binits{M.A.}},
\bauthor{\bsnm{Salwana}, \binits{E.}},
\bauthor{\bsnm{Zulkifli}, \binits{Z.}},
\bauthor{\bsnm{Nayan}, \binits{N.M.}},
\bauthor{\bsnm{Abdullah}, \binits{N.}}:
\bctitle{Prediction of learning disorder: A-systematic review}.
In: \bbtitle{International Visual Informatics Conference},
pp. \bfpage{429}--\blpage{440}
(\byear{2019}).
\bcomment{Springer}
\end{bchapter}
\endbibitem

\bibitem{hamstra1993longitudinal}
\begin{barticle}
\bauthor{\bsnm{Hamstra-Bletz}, \binits{L.}},
\bauthor{\bsnm{Bl{\"o}te}, \binits{A.W.}}:
\batitle{A longitudinal study on dysgraphic handwriting in primary school}.
\bjtitle{Journal of learning disabilities}
\bvolume{26}(\bissue{10}),
\bfpage{689}--\blpage{699}
(\byear{1993})
\end{barticle}
\endbibitem

\bibitem{brown1981learning}
\begin{botherref}
\oauthor{\bsnm{Brown}, \binits{J.}}:
Learning disabilities: A paediatric neurologist’s point of view.
Transactions of the College of Medicine of South Africa
(December),
49--104
(1981)
\end{botherref}
\endbibitem

\bibitem{rosenblum2004handwriting}
\begin{barticle}
\bauthor{\bsnm{Rosenblum}, \binits{S.}},
\bauthor{\bsnm{Weiss}, \binits{P.L.}},
\bauthor{\bsnm{Parush}, \binits{S.}}:
\batitle{Handwriting evaluation for developmental dysgraphia: Process versus
  product}.
\bjtitle{Reading and writing}
\bvolume{17}(\bissue{5}),
\bfpage{433}--\blpage{458}
(\byear{2004})
\end{barticle}
\endbibitem

\bibitem{beery2004beery}
\begin{botherref}
\oauthor{\bsnm{Beery}, \binits{K.E.}}:
Beery vmi: The beery-buktenica developmental test of visual-motor integration.
Minneapolis, MN: Pearson
(2004)
\end{botherref}
\endbibitem

\bibitem{mehrinejad2012investigation}
\begin{botherref}
\oauthor{\bsnm{Mehrinejad}, \binits{S.}},
\oauthor{\bsnm{SOBHI}, \binits{G.N.}},
\oauthor{\bsnm{RAJABI}, \binits{M.S.}}:
An investigation of the power of the bender gestalt test in the prediction of
  preschool children's predisposition for dyslexia and dysgraphia
(2012)
\end{botherref}
\endbibitem

\bibitem{korkman2014nepsy}
\begin{bbook}
\bauthor{\bsnm{Korkman}, \binits{M.}},
\bauthor{\bsnm{Kirk}, \binits{U.}},
\bauthor{\bsnm{Kemp}, \binits{S.}}:
\bbtitle{NEPSY-ii}.
\bpublisher{Pearson Madrid}, \blocation{???}
(\byear{2014})
\end{bbook}
\endbibitem

\bibitem{chung2015dysgraphia}
\begin{barticle}
\bauthor{\bsnm{Chung}, \binits{P.}},
\bauthor{\bsnm{Patel}, \binits{D.R.}}:
\batitle{Dysgraphia}.
\bjtitle{International Journal of Child and Adolescent Health}
\bvolume{8}(\bissue{1}),
\bfpage{27}
(\byear{2015})
\end{barticle}
\endbibitem

\bibitem{meyers1995rey}
\begin{barticle}
\bauthor{\bsnm{Meyers}, \binits{J.E.}},
\bauthor{\bsnm{Meyers}, \binits{K.R.}}:
\batitle{Rey complex figure test under four different administration
  procedures}.
\bjtitle{The Clinical Neuropsychologist}
\bvolume{9}(\bissue{1}),
\bfpage{63}--\blpage{67}
(\byear{1995})
\end{barticle}
\endbibitem

\bibitem{roth2014assessment}
\begin{bchapter}
\bauthor{\bsnm{Roth}, \binits{R.M.}},
\bauthor{\bsnm{Isquith}, \binits{P.K.}},
\bauthor{\bsnm{Gioia}, \binits{G.A.}}:
\bctitle{Assessment of executive functioning using the behavior rating
  inventory of executive function (brief)}.
In: \bbtitle{Handbook of Executive Functioning},
pp. \bfpage{301}--\blpage{331}.
\bpublisher{Springer}, \blocation{???}
(\byear{2014})
\end{bchapter}
\endbibitem

\bibitem{mccloskey2017developmental}
\begin{barticle}
\bauthor{\bsnm{McCloskey}, \binits{M.}},
\bauthor{\bsnm{Rapp}, \binits{B.}}:
\batitle{Developmental dysgraphia: An overview and framework for research}.
\bjtitle{Cognitive neuropsychology}
\bvolume{34}(\bissue{3-4}),
\bfpage{65}--\blpage{82}
(\byear{2017})
\end{barticle}
\endbibitem

\bibitem{burns2010wechsler}
\begin{barticle}
\bauthor{\bsnm{Burns}, \binits{T.G.}}:
\batitle{Wechsler individual achievement test-iii: What is the ‘gold
  standard’for measuring academic achievement?}
\bjtitle{Applied Neuropsychology}
\bvolume{17}(\bissue{3}),
\bfpage{234}--\blpage{236}
(\byear{2010})
\end{barticle}
\endbibitem

\bibitem{woodcock2007woodcock}
\begin{botherref}
\oauthor{\bsnm{Woodcock}, \binits{R.}},
\oauthor{\bsnm{McGrew}, \binits{K.}},
\oauthor{\bsnm{Mather}, \binits{N.}},
\oauthor{\bsnm{Schrank}, \binits{F.}}:
Woodcock-johnson iii nu tests of achievement
(2007)
\end{botherref}
\endbibitem

\bibitem{hammill2009test}
\begin{bbook}
\bauthor{\bsnm{Hammill}, \binits{D.D.}},
\bauthor{\bsnm{Larsen}, \binits{S.C.}}:
\bbtitle{Test of Written Language: TOWL4}.
\bpublisher{Pro-ed}, \blocation{???}
(\byear{2009})
\end{bbook}
\endbibitem

\bibitem{wagner1999comprehensive}
\begin{bbook}
\bauthor{\bsnm{Wagner}, \binits{R.K.}},
\bauthor{\bsnm{Torgesen}, \binits{J.K.}},
\bauthor{\bsnm{Rashotte}, \binits{C.A.}},
\bauthor{\bsnm{Pearson}, \binits{N.A.}}:
\bbtitle{Comprehensive Test of Phonological Processing: CTOPP}.
\bpublisher{Pro-ed Austin, TX}, \blocation{???}
(\byear{1999})
\end{bbook}
\endbibitem

\bibitem{crouch2007dysgraphia}
\begin{barticle}
\bauthor{\bsnm{Crouch}, \binits{A.L.}},
\bauthor{\bsnm{Jakubecy}, \binits{J.J.}}:
\batitle{Dysgraphia: How it affects a student's performance and what can be
  done about it.}
\bjtitle{TEACHING Exceptional Children Plus}
\bvolume{3}(\bissue{3}),
\bfpage{3}
(\byear{2007})
\end{barticle}
\endbibitem

\bibitem{reynolds2007test}
\begin{bbook}
\bauthor{\bsnm{Reynolds}, \binits{C.R.}},
\bauthor{\bsnm{Voress}, \binits{J.K.}}:
\bbtitle{Test of Memory and Learning (TOMAL 2)}.
\bpublisher{Pro-Ed Austin, TX}, \blocation{???}
(\byear{2007})
\end{bbook}
\endbibitem

\bibitem{hartman2007wide}
\begin{botherref}
\oauthor{\bsnm{Hartman}, \binits{D.E.}}:
Wide range assessment of memory and learning-2 (wraml-2): Wredesigned and
  wreally improved
(2007)
\end{botherref}
\endbibitem

\bibitem{petermann2011wechsler}
\begin{bbook}
\bauthor{\bsnm{Petermann}, \binits{F.}}:
\bbtitle{Wechsler Intelligence Scale for children:(WISC-IV)}.
\bpublisher{Pearson}, \blocation{???}
(\byear{2011})
\end{bbook}
\endbibitem

\bibitem{elliott1990differential}
\begin{botherref}
\oauthor{\bsnm{Elliott}, \binits{C.D.}},
\oauthor{\bsnm{Murray}, \binits{G.}},
\oauthor{\bsnm{Pearson}, \binits{L.}}:
Differential ability scales.
San Antonio, Texas
(1990)
\end{botherref}
\endbibitem

\bibitem{kushki2011changes}
\begin{barticle}
\bauthor{\bsnm{Kushki}, \binits{A.}},
\bauthor{\bsnm{Schwellnus}, \binits{H.}},
\bauthor{\bsnm{Ilyas}, \binits{F.}},
\bauthor{\bsnm{Chau}, \binits{T.}}:
\batitle{Changes in kinetics and kinematics of handwriting during a prolonged
  writing task in children with and without dysgraphia}.
\bjtitle{Research in developmental disabilities}
\bvolume{32}(\bissue{3}),
\bfpage{1058}--\blpage{1064}
(\byear{2011})
\end{barticle}
\endbibitem

\bibitem{rosenblum2003air}
\begin{barticle}
\bauthor{\bsnm{Rosenblum}, \binits{S.}},
\bauthor{\bsnm{Parush}, \binits{S.}},
\bauthor{\bsnm{Weiss}, \binits{P.L.}}:
\batitle{The in air phenomenon: Temporal and spatial correlates of the
  handwriting process}.
\bjtitle{Perceptual and Motor skills}
\bvolume{96}(\bissue{3}),
\bfpage{933}--\blpage{954}
(\byear{2003})
\end{barticle}
\endbibitem

\bibitem{Mekyska2017}
\begin{barticle}
\bauthor{\bsnm{Mekyska}, \binits{J.}},
\bauthor{\bsnm{Faundez-Zanuy}, \binits{M.}},
\bauthor{\bsnm{Mzourek}, \binits{Z.}},
\bauthor{\bsnm{Galaz}, \binits{Z.}},
\bauthor{\bsnm{Smekal}, \binits{Z.}},
\bauthor{\bsnm{Rosenblum}, \binits{S.}}:
\batitle{{Identification and Rating of Developmental Dysgraphia by Handwriting
  Analysis}}.
\bjtitle{IEEE Transactions on Human-Machine Systems}
\bvolume{47}(\bissue{2}),
\bfpage{235}--\blpage{248}
(\byear{2017}).
\doiurl{10.1109/THMS.2016.2586605}
\end{barticle}
\endbibitem

\bibitem{Asselborn2018}
\begin{botherref}
\oauthor{\bsnm{Asselborn}, \binits{T.}},
\oauthor{\bsnm{Gargot}, \binits{T.}},
\oauthor{\bsnm{Kidzi{\'{n}}ski}, \binits{{\L}.}},
\oauthor{\bsnm{Johal}, \binits{W.}},
\oauthor{\bsnm{Cohen}, \binits{D.}},
\oauthor{\bsnm{Jolly}, \binits{C.}},
\oauthor{\bsnm{Dillenbourg}, \binits{P.}}:
{Automated human-level diagnosis of dysgraphia using a consumer tablet}.
npj Digital Medicine
\textbf{1}(1)
(2018).
\doiurl{10.1038/s41746-018-0049-x}
\end{botherref}
\endbibitem

\bibitem{Drotar2020}
\begin{barticle}
\bauthor{\bsnm{Drot{\'{a}}r}, \binits{P.}},
\bauthor{\bsnm{Dobe{\v{s}}}, \binits{M.}}:
\batitle{{Dysgraphia detection through machine learning}}.
\bjtitle{Scientific Reports}
\bvolume{10}(\bissue{1}),
\bfpage{1}--\blpage{11}
(\byear{2020}).
\doiurl{10.1038/s41598-020-78611-9}
\end{barticle}
\endbibitem

\bibitem{Asselborn2020}
\begin{barticle}
\bauthor{\bsnm{Asselborn}, \binits{T.}},
\bauthor{\bsnm{Chapatte}, \binits{M.}},
\bauthor{\bsnm{Dillenbourg}, \binits{P.}}:
\batitle{{Extending the Spectrum of Dysgraphia: A Data Driven Strategy to
  Estimate Handwriting Quality}}.
\bjtitle{Scientific Reports}
\bvolume{10}(\bissue{1}),
\bfpage{1}--\blpage{11}
(\byear{2020}).
\doiurl{10.1038/s41598-020-60011-8}
\end{barticle}
\endbibitem

\bibitem{Devillaine2021}
\begin{botherref}
\oauthor{\bsnm{Devillaine}, \binits{L.}},
\oauthor{\bsnm{Lambert}, \binits{R.}},
\oauthor{\bsnm{Boutet}, \binits{J.}},
\oauthor{\bsnm{Aloui}, \binits{S.}},
\oauthor{\bsnm{Brault}, \binits{V.}},
\oauthor{\bsnm{Jolly}, \binits{C.}},
\oauthor{\bsnm{Labyt}, \binits{E.}}:
{Analysis of graphomotor tests with machine learning algorithms for an early
  and universal pre-diagnosis of dysgraphia}.
Sensors
\textbf{21}(21)
(2021).
\doiurl{10.3390/s21217026}
\end{botherref}
\endbibitem

\bibitem{Deschamps2021}
\begin{botherref}
\oauthor{\bsnm{Deschamps}, \binits{L.}},
\oauthor{\bsnm{Devillaine}, \binits{L.}},
\oauthor{\bsnm{Gaffet}, \binits{C.}},
\oauthor{\bsnm{Lambert}, \binits{R.}},
\oauthor{\bsnm{Aloui}, \binits{S.}},
\oauthor{\bsnm{Boutet}, \binits{J.}},
\oauthor{\bsnm{Brault}, \binits{V.}},
\oauthor{\bsnm{Labyt}, \binits{E.}},
\oauthor{\bsnm{Jolly}, \binits{C.}},
\oauthor{\bparticle{al} \bsnm{De}}:
{Development of a Pre-Diagnosis Tool Based on Machine Learning Algorithms on
  the BHK Test to Improve the Diagnosis of Dysgraphia}.
Advances in Artificial Intelligence and Machine Learning,
222--13194
(2021)
\end{botherref}
\endbibitem

\bibitem{Dankovicova2019}
\begin{bchapter}
\bauthor{\bsnm{Dankovicova}, \binits{Z.}},
\bauthor{\bsnm{Hurtuk}, \binits{J.}},
\bauthor{\bsnm{Fecilak}, \binits{P.}}:
\bctitle{{Evaluation of digitalized handwriting for dysgraphia detection using
  random forest classification method}}.
In: \bbtitle{SISY 2019 - IEEE 17th International Symposium on Intelligent
  Systems and Informatics, Proceedings}
(\byear{2019}).
\doiurl{10.1109/SISY47553.2019.9111567}
\end{bchapter}
\endbibitem

\bibitem{rosenblum2016identifying}
\begin{barticle}
\bauthor{\bsnm{Rosenblum}, \binits{S.}},
\bauthor{\bsnm{Dror}, \binits{G.}}:
\batitle{Identifying developmental dysgraphia characteristics utilizing
  handwriting classification methods}.
\bjtitle{IEEE Transactions on Human-Machine Systems}
\bvolume{47}(\bissue{2}),
\bfpage{293}--\blpage{298}
(\byear{2016})
\end{barticle}
\endbibitem

\bibitem{Mekyska2019}
\begin{botherref}
\oauthor{\bsnm{Mekyska}, \binits{J.}},
\oauthor{\bsnm{Bednarova}, \binits{J.}},
\oauthor{\bsnm{Faundez-Zanuy}, \binits{M.}},
\oauthor{\bsnm{Galaz}, \binits{Z.}},
\oauthor{\bsnm{Safarova}, \binits{K.}},
\oauthor{\bsnm{Zvoncak}, \binits{V.}},
\oauthor{\bsnm{Mucha}, \binits{J.}},
\oauthor{\bsnm{Smekal}, \binits{Z.}},
\oauthor{\bsnm{Ondrackova}, \binits{A.}},
\oauthor{\bsnm{Urbanek}, \binits{T.}},
\oauthor{\bsnm{Havigerova}, \binits{J.M.}}:
{Computerised Assessment of Graphomotor Difficulties in a Cohort of School-aged
  Children}.
International Congress on Ultra Modern Telecommunications and Control Systems
  and Workshops
\textbf{2019-Octob}
(2019).
\doiurl{10.1109/ICUMT48472.2019.8970767}
\end{botherref}
\endbibitem

\bibitem{devi2021early}
\begin{bchapter}
\bauthor{\bsnm{Devi}, \binits{A.}},
\bauthor{\bsnm{Kavya}, \binits{G.}},
\bauthor{\bsnm{Therese}, \binits{M.J.}},
\bauthor{\bsnm{Gayathri}, \binits{R.}}:
\bctitle{Early diagnosing and identifying tool for specific learning disability
  using decision tree algorithm}.
In: \bbtitle{2021 Third International Conference on Inventive Research in
  Computing Applications (ICIRCA)},
pp. \bfpage{1445}--\blpage{1450}
(\byear{2021}).
\bcomment{IEEE}
\end{bchapter}
\endbibitem

\bibitem{kedar2021identifying}
\begin{barticle}
\bauthor{\bsnm{Kedar}, \binits{S.}}, \betal:
\batitle{Identifying learning disability through digital handwriting analysis}.
\bjtitle{Turkish Journal of Computer and Mathematics Education (TURCOMAT)}
\bvolume{12}(\bissue{1S}),
\bfpage{46}--\blpage{56}
(\byear{2021})
\end{barticle}
\endbibitem

\bibitem{Biau2012}
\begin{botherref}
\oauthor{\bsnm{Biau}, \binits{G.}}:
{Analysis of a random forests model}
(2012)
\end{botherref}
\endbibitem

\bibitem{Izenman2013}
\begin{botherref}
\oauthor{\bsnm{Izenman}, \binits{A.J.}}:
{Linear Discriminant Analysis},
237--280
(2013).
\doiurl{10.1007/978-0-387-78189-1_8}
\end{botherref}
\endbibitem

\bibitem{LaerdStatistics2015}
\begin{botherref}
\oauthor{\bsnm{{Laerd Statistics}}}:
{Mann-Whitney U test using SPSS Statistics.}
Statistical tutorials and software guides.
(2015)
\end{botherref}
\endbibitem

\bibitem{Chen2016}
\begin{bchapter}
\bauthor{\bsnm{Chen}, \binits{T.}},
\bauthor{\bsnm{Guestrin}, \binits{C.}}:
\bctitle{{XGBoost: A scalable tree boosting system}}.
In: \bbtitle{Proceedings of the ACM SIGKDD International Conference on
  Knowledge Discovery and Data Mining},
vol. \bseriesno{13-17-August-2016}
(\byear{2016}).
\doiurl{10.1145/2939672.2939785}
\end{bchapter}
\endbibitem

\bibitem{Dui2020}
\begin{barticle}
\bauthor{\bsnm{Dui}, \binits{L.G.}},
\bauthor{\bsnm{Lunardini}, \binits{F.}},
\bauthor{\bsnm{Termine}, \binits{C.}},
\bauthor{\bsnm{Matteucci}, \binits{M.}},
\bauthor{\bsnm{Ferrante}, \binits{S.}}:
\batitle{{A Tablet-Based App to Discriminate Children at Potential Risk of
  Handwriting Alterations in a Preliteracy Stage}}.
\bjtitle{Proceedings of the Annual International Conference of the IEEE
  Engineering in Medicine and Biology Society, EMBS}
\bvolume{2020-July},
\bfpage{5856}--\blpage{5859}
(\byear{2020}).
\doiurl{10.1109/EMBC44109.2020.9176041}
\end{barticle}
\endbibitem

\bibitem{Peng2002}
\begin{botherref}
\oauthor{\bsnm{Peng}, \binits{C.Y.J.}},
\oauthor{\bsnm{Lee}, \binits{K.L.}},
\oauthor{\bsnm{Ingersoll}, \binits{G.M.}}:
{An introduction to logistic regression analysis and reporting}.
Journal of Educational Research
\textbf{96}(1)
(2002).
\doiurl{10.1080/00220670209598786}
\end{botherref}
\endbibitem

\bibitem{Zizka2019}
\begin{bchapter}
\bauthor{\bsnm{{\v{Z}}i{\v{z}}ka}, \binits{J.}},
\bauthor{\bsnm{Dařena}, \binits{F.}},
\bauthor{\bsnm{Svoboda}, \binits{A.}}:
\bctitle{{Adaboost}}.
In: \bbtitle{Text Mining with Machine Learning},
(\byear{2019}).
\doiurl{10.1201/9780429469275-9}
\end{bchapter}
\endbibitem

\bibitem{Richard2020}
\begin{botherref}
\oauthor{\bsnm{Richard}, \binits{G.}},
\oauthor{\bsnm{Serrurier}, \binits{M.}}:
{Dyslexia and Dysgraphia prediction: A new machine learning approach}.
arXiv
(2020)
\end{botherref}
\endbibitem

\bibitem{Burges1998}
\begin{botherref}
\oauthor{\bsnm{Burges}, \binits{C.J.C.}}:
{A tutorial on support vector machines for pattern recognition}.
Data Mining and Knowledge Discovery
\textbf{2}(2)
(1998).
\doiurl{10.1023/A:1009715923555}
\end{botherref}
\endbibitem

\bibitem{Yegnanarayana1994}
\begin{botherref}
\oauthor{\bsnm{Yegnanarayana}, \binits{B.}}:
{Artificial neural networks for pattern recognition}.
Sadhana
\textbf{19}(2)
(1994).
\doiurl{10.1007/BF02811896}
\end{botherref}
\endbibitem

\bibitem{Wu2008}
\begin{botherref}
\oauthor{\bsnm{Wu}, \binits{X.}},
\oauthor{\bsnm{Kumar}, \binits{V.}},
\oauthor{\bsnm{Ross}, \binits{Q.J.}},
\oauthor{\bsnm{Ghosh}, \binits{J.}},
\oauthor{\bsnm{Yang}, \binits{Q.}},
\oauthor{\bsnm{Motoda}, \binits{H.}},
\oauthor{\bsnm{McLachlan}, \binits{G.J.}},
\oauthor{\bsnm{Ng}, \binits{A.}},
\oauthor{\bsnm{Liu}, \binits{B.}},
\oauthor{\bsnm{Yu}, \binits{P.S.}},
\oauthor{\bsnm{Zhou}, \binits{Z.H.}},
\oauthor{\bsnm{Steinbach}, \binits{M.}},
\oauthor{\bsnm{Hand}, \binits{D.J.}},
\oauthor{\bsnm{Steinberg}, \binits{D.}}:
{Top 10 algorithms in data mining}.
Knowledge and Information Systems
\textbf{14}(1)
(2008).
\doiurl{10.1007/s10115-007-0114-2}
\end{botherref}
\endbibitem

\bibitem{Et.al.2021}
\begin{botherref}
\oauthor{\bsnm{Et.~al.}, \binits{S.D.}}:
{Comparison of Classification Methods used in Machine Learning for Dysgraphia
  Identification}.
Turkish Journal of Computer and Mathematics Education (TURCOMAT)
\textbf{12}(11)
(2021).
\doiurl{10.17762/turcomat.v12i11.6142}
\end{botherref}
\endbibitem

\bibitem{Jenul2021}
\begin{botherref}
\oauthor{\bsnm{Jenul}, \binits{A.}},
\oauthor{\bsnm{Schrunner}, \binits{S.}},
\oauthor{\bsnm{Liland}, \binits{K.H.}},
\oauthor{\bsnm{Indahl}, \binits{U.G.}},
\oauthor{\bsnm{Futsaether}, \binits{C.M.}},
\oauthor{\bsnm{Tomic}, \binits{O.}}:
{Rent - Repeated elastic net technique for feature selection}.
IEEE Access
\textbf{9}
(2021).
\doiurl{10.1109/ACCESS.2021.3126429}
\end{botherref}
\endbibitem

\bibitem{Zhang2016}
\begin{botherref}
\oauthor{\bsnm{Zhang}, \binits{Z.}}:
{Introduction to machine learning: K-nearest neighbors}.
Annals of Translational Medicine
\textbf{4}(11)
(2016).
\doiurl{10.21037/atm.2016.03.37}
\end{botherref}
\endbibitem

\bibitem{Quinlan1986}
\begin{botherref}
\oauthor{\bsnm{Quinlan}, \binits{J.R.}}:
{Induction of Decision Trees}.
Machine Learning
\textbf{1}(1)
(1986).
\doiurl{10.1023/A:1022643204877}
\end{botherref}
\endbibitem

\bibitem{Hewapathirana2021}
\begin{bchapter}
\bauthor{\bsnm{Hewapathirana}, \binits{C.}},
\bauthor{\bsnm{Abeysinghe}, \binits{K.}},
\bauthor{\bsnm{Maheshani}, \binits{P.}},
\bauthor{\bsnm{Liyanage}, \binits{P.}},
\bauthor{\bsnm{Krishara}, \binits{J.}},
\bauthor{\bsnm{Thelijjagoda}, \binits{S.}}:
\bctitle{{A Mobile-Based Screening and Refinement System to Identify the Risk
  of Dyscalculia and Dysgraphia Learning Disabilities in Primary School
  Students}}.
In: \bbtitle{2021 10th International Conference on Information and Automation
  for Sustainability, ICIAfS 2021}
(\byear{2021}).
\doiurl{10.1109/ICIAfS52090.2021.9605998}
\end{bchapter}
\endbibitem

\bibitem{Sihwi2019}
\begin{bchapter}
\bauthor{\bsnm{Sihwi}, \binits{S.W.}},
\bauthor{\bsnm{Fikri}, \binits{K.}},
\bauthor{\bsnm{Aziz}, \binits{A.}}:
\bctitle{{Dysgraphia Identification from Handwriting with Support Vector
  Machine Method}}.
In: \bbtitle{Journal of Physics: Conference Series},
vol. \bseriesno{1201}
(\byear{2019}).
\doiurl{10.1088/1742-6596/1201/1/012050}
\end{bchapter}
\endbibitem

\bibitem{Kurniawan2018}
\begin{bchapter}
\bauthor{\bsnm{Kurniawan}, \binits{D.A.}},
\bauthor{\bsnm{Sihwi}, \binits{S.W.}},
\bauthor{\bsnm{Gunarhadi}}:
\bctitle{{An expert system for diagnosing dysgraphia}}.
In: \bbtitle{Proceedings - 2017 2nd International Conferences on Information
  Technology, Information Systems and Electrical Engineering, ICITISEE 2017},
vol. \bseriesno{2018-January}
(\byear{2018}).
\doiurl{10.1109/ICITISEE.2017.8285552}
\end{bchapter}
\endbibitem

\bibitem{zvoncak2018effect}
\begin{bchapter}
\bauthor{\bsnm{Zvoncak}, \binits{V.}},
\bauthor{\bsnm{Mekyska}, \binits{J.}},
\bauthor{\bsnm{Safarova}, \binits{K.}},
\bauthor{\bsnm{Galaz}, \binits{Z.}},
\bauthor{\bsnm{Mucha}, \binits{J.}},
\bauthor{\bsnm{Kiska}, \binits{T.}},
\bauthor{\bsnm{Smekal}, \binits{Z.}},
\bauthor{\bsnm{Losenicka}, \binits{B.}},
\bauthor{\bsnm{Cechova}, \binits{B.}},
\bauthor{\bsnm{Francova}, \binits{P.}}, \betal:
\bctitle{Effect of stroke-level intra-writer normalization on computerized
  assessment of developmental dysgraphia}.
In: \bbtitle{2018 10th International Congress on Ultra Modern
  Telecommunications and Control Systems and Workshops (ICUMT)},
pp. \bfpage{1}--\blpage{5}
(\byear{2018}).
\bcomment{IEEE}
\end{bchapter}
\endbibitem

\bibitem{kariyawasam2019pubudu}
\begin{bchapter}
\bauthor{\bsnm{Kariyawasam}, \binits{R.}},
\bauthor{\bsnm{Nadeeshani}, \binits{M.}},
\bauthor{\bsnm{Hamid}, \binits{T.}},
\bauthor{\bsnm{Subasinghe}, \binits{I.}},
\bauthor{\bsnm{Samarasinghe}, \binits{P.}},
\bauthor{\bsnm{Ratnayake}, \binits{P.}}:
\bctitle{Pubudu: Deep learning based screening and intervention of dyslexia,
  dysgraphia and dyscalculia}.
In: \bbtitle{2019 14th Conference on Industrial and Information Systems
  (ICIIS)},
pp. \bfpage{476}--\blpage{481}
(\byear{2019}).
\bcomment{IEEE}
\end{bchapter}
\endbibitem

\bibitem{yogarajah2020deep}
\begin{bchapter}
\bauthor{\bsnm{Yogarajah}, \binits{P.}},
\bauthor{\bsnm{Bhushan}, \binits{B.}}:
\bctitle{Deep learning approach to automated detection of dyslexia-dysgraphia}.
In: \bbtitle{The 25th IEEE International Conference on Pattern Recognition}
(\byear{2020})
\end{bchapter}
\endbibitem

\bibitem{Isa2019}
\begin{barticle}
\bauthor{\bsnm{Isa}, \binits{I.S.}},
\bauthor{\bsnm{{Syazwani Rahimi}}, \binits{W.N.}},
\bauthor{\bsnm{Ramlan}, \binits{S.A.}},
\bauthor{\bsnm{Sulaiman}, \binits{S.N.}}:
\batitle{{Automated Detection of Dyslexia Symptom Based on Handwriting Image
  for Primary School Children}}.
\bjtitle{Procedia Computer Science}
\bvolume{163},
\bfpage{440}--\blpage{449}
(\byear{2019}).
\doiurl{10.1016/j.procs.2019.12.127}
\end{barticle}
\endbibitem

\bibitem{Zvoncak2019}
\begin{botherref}
\oauthor{\bsnm{Zvoncak}, \binits{V.}},
\oauthor{\bsnm{Mekyska}, \binits{J.}},
\oauthor{\bsnm{Safarova}, \binits{K.}},
\oauthor{\bsnm{Smekal}, \binits{Z.}},
\oauthor{\bsnm{Brezany}, \binits{P.}}:
{New approach of dysgraphic handwriting analysis based on the tunable Q-factor
  wavelet transform}.
2019 42nd International Convention on Information and Communication Technology,
  Electronics and Microelectronics, MIPRO 2019 - Proceedings,
289--294
(2019).
\doiurl{10.23919/MIPRO.2019.8756872}
\end{botherref}
\endbibitem

\bibitem{dui2020tablet}
\begin{barticle}
\bauthor{\bsnm{Dui}, \binits{L.G.}},
\bauthor{\bsnm{Lunardini}, \binits{F.}},
\bauthor{\bsnm{Termine}, \binits{C.}},
\bauthor{\bsnm{Matteucci}, \binits{M.}},
\bauthor{\bsnm{Stucchi}, \binits{N.A.}},
\bauthor{\bsnm{Borghese}, \binits{N.A.}},
\bauthor{\bsnm{Ferrante}, \binits{S.}}:
\batitle{A tablet app for handwriting skill screening at the preliteracy stage:
  Instrument validation study}.
\bjtitle{JMIR serious games}
\bvolume{8}(\bissue{4}),
\bfpage{20126}
(\byear{2020})
\end{barticle}
\endbibitem

\bibitem{viviani1982trajectory}
\begin{barticle}
\bauthor{\bsnm{Viviani}, \binits{P.}},
\bauthor{\bsnm{Terzuolo}, \binits{C.}}:
\batitle{Trajectory determines movement dynamics}.
\bjtitle{Neuroscience}
\bvolume{7}(\bissue{2}),
\bfpage{431}--\blpage{437}
(\byear{1982})
\end{barticle}
\endbibitem

\bibitem{lashley1951problem}
\begin{bbook}
\bauthor{\bsnm{Lashley}, \binits{K.S.}}, \betal:
\bbtitle{The Problem of Serial Order in Behavior}
vol. \bseriesno{21}.
\bpublisher{Bobbs-Merrill Oxford, United Kingdom}, \blocation{???}
(\byear{1951})
\end{bbook}
\endbibitem

\bibitem{giordano2014addressing}
\begin{bchapter}
\bauthor{\bsnm{Giordano}, \binits{D.}},
\bauthor{\bsnm{Maiorana}, \binits{F.}}:
\bctitle{Addressing dysgraphia with a mobile, web-based software with
  interactive feedback}.
In: \bbtitle{IEEE-EMBS International Conference on Biomedical and Health
  Informatics (BHI)},
pp. \bfpage{264}--\blpage{268}
(\byear{2014}).
\bcomment{IEEE}
\end{bchapter}
\endbibitem

\bibitem{raza2017interactive}
\begin{bchapter}
\bauthor{\bsnm{Raza}, \binits{T.F.}},
\bauthor{\bsnm{Arif}, \binits{H.}},
\bauthor{\bsnm{Darvagheh}, \binits{S.H.}},
\bauthor{\bsnm{Hajjdiab}, \binits{H.}}:
\bctitle{Interactive mobile application for testing children with dysgraphia}.
In: \bbtitle{Proceedings of the 9th International Conference on Machine
  Learning and Computing},
pp. \bfpage{432}--\blpage{436}
(\byear{2017})
\end{bchapter}
\endbibitem

\bibitem{Dimauro2020}
\begin{barticle}
\bauthor{\bsnm{Dimauro}, \binits{G.}},
\bauthor{\bsnm{Bevilacqua}, \binits{V.}},
\bauthor{\bsnm{Colizzi}, \binits{L.}},
\bauthor{\bsnm{{Di Pierro}}, \binits{D.}}:
\batitle{{TestGraphia, a software system for the early diagnosis of
  dysgraphia}}.
\bjtitle{IEEE Access}
\bvolume{8},
\bfpage{19564}--\blpage{19575}
(\byear{2020}).
\doiurl{10.1109/ACCESS.2020.2968367}
\end{barticle}
\endbibitem

\bibitem{Le}
\begin{botherref}
Lexcercise:dysgrphia-test.
\url{https://www.lexercise.com/tests/dysgraphia-test}.
Accessed: 2022-02-27
\end{botherref}
\endbibitem

\bibitem{Addit}
\begin{botherref}
Additude:dysgrphia-test.
\url{https://www.additudemag.com/screener-dysgraphia-symptoms-test-children/?src=embed_link}.
Accessed: 2022-02-27
\end{botherref}
\endbibitem

\bibitem{Dy}
\begin{botherref}
Dyscreen.
\url{https://dystech.com.au/}.
Accessed: 2022-02-27
\end{botherref}
\endbibitem

\bibitem{AIdy}
\begin{botherref}
AI Dysgraphia Pre-screening.
\url{https://appadvice.com/app/ai-dysgraphia-pre-screening/1546707440}.
Accessed: 2022-02-27
\end{botherref}
\endbibitem

\bibitem{wu2019automated}
\begin{barticle}
\bauthor{\bsnm{Wu}, \binits{Z.}},
\bauthor{\bsnm{Lin}, \binits{T.}},
\bauthor{\bsnm{Li}, \binits{M.}}:
\batitle{Automated detection of children at risk of chinese handwriting
  difficulties using handwriting process information: An exploratory study}.
\bjtitle{IEICE TRANSACTIONS on Information and Systems}
\bvolume{102}(\bissue{1}),
\bfpage{147}--\blpage{155}
(\byear{2019})
\end{barticle}
\endbibitem

\bibitem{hen2019characteristics}
\begin{barticle}
\bauthor{\bsnm{Hen-Herbst}, \binits{L.}},
\bauthor{\bsnm{Rosenblum}, \binits{S.}}:
\batitle{Which characteristics predict writing capabilities among adolescents
  with dysgraphia?}
\bjtitle{Pattern Recognition Letters}
\bvolume{121},
\bfpage{6}--\blpage{12}
(\byear{2019})
\end{barticle}
\endbibitem

\bibitem{Guilbert2019}
\begin{barticle}
\bauthor{\bsnm{Guilbert}, \binits{J.}},
\bauthor{\bsnm{Alamargot}, \binits{D.}},
\bauthor{\bsnm{Morin}, \binits{M.F.}}:
\batitle{{Handwriting on a tablet screen: Role of visual and proprioceptive
  feedback in the control of movement by children and adults}}.
\bjtitle{Human Movement Science}
(\byear{2019}).
\doiurl{10.1016/j.humov.2018.09.001}
\end{barticle}
\endbibitem

\bibitem{Prunty2020}
\begin{barticle}
\bauthor{\bsnm{Prunty}, \binits{M.M.}},
\bauthor{\bsnm{Pratt}, \binits{A.}},
\bauthor{\bsnm{Raman}, \binits{E.}},
\bauthor{\bsnm{Simmons}, \binits{L.}},
\bauthor{\bsnm{Steele-Bobat}, \binits{F.}}:
\batitle{{Grip strength and pen pressure are not key contributors to
  handwriting difficulties in children with developmental coordination
  disorder}}.
\bjtitle{British Journal of Occupational Therapy}
\bvolume{83}(\bissue{6}),
\bfpage{387}--\blpage{396}
(\byear{2020}).
\doiurl{10.1177/0308022619885046}
\end{barticle}
\endbibitem

\bibitem{Lin2017}
\begin{barticle}
\bauthor{\bsnm{Lin}, \binits{Y.C.}},
\bauthor{\bsnm{Chao}, \binits{Y.L.}},
\bauthor{\bsnm{Wu}, \binits{S.K.}},
\bauthor{\bsnm{Lin}, \binits{H.H.}},
\bauthor{\bsnm{Hsu}, \binits{C.H.}},
\bauthor{\bsnm{Hsu}, \binits{H.M.}},
\bauthor{\bsnm{Kuo}, \binits{L.C.}}:
\batitle{{Comprehension of handwriting development: Pen-grip kinetics in
  handwriting tasks and its relation to fine motor skills among school-age
  children}}.
\bjtitle{Australian Occupational Therapy Journal}
\bvolume{64}(\bissue{5}),
\bfpage{369}--\blpage{380}
(\byear{2017}).
\doiurl{10.1111/1440-1630.12393}
\end{barticle}
\endbibitem

\bibitem{Biotteau2019}
\begin{barticle}
\bauthor{\bsnm{Biotteau}, \binits{M.}},
\bauthor{\bsnm{Danna}, \binits{J.}},
\bauthor{\bsnm{Baudou}, \binits{{\'{E}}.}},
\bauthor{\bsnm{Puyjarinet}, \binits{F.}},
\bauthor{\bsnm{Velay}, \binits{J.L.}},
\bauthor{\bsnm{Albaret}, \binits{J.M.}},
\bauthor{\bsnm{Chaix}, \binits{Y.}}:
\batitle{{Developmental coordination disorder and dysgraphia: Signs and
  symptoms, diagnosis, and rehabilitation}}.
\bjtitle{Neuropsychiatric Disease and Treatment}
\bvolume{15},
\bfpage{1873}--\blpage{1885}
(\byear{2019}).
\doiurl{10.2147/NDT.S120514}
\end{barticle}
\endbibitem

\bibitem{herrick1961pressure}
\begin{barticle}
\bauthor{\bsnm{Herrick}, \binits{V.E.}},
\bauthor{\bsnm{Otto}, \binits{W.}}:
\batitle{Pressure on point and barrel of a writing instrument}.
\bjtitle{The Journal of Experimental Education}
\bvolume{30}(\bissue{2}),
\bfpage{215}--\blpage{230}
(\byear{1961})
\end{barticle}
\endbibitem

\bibitem{falk2010grip}
\begin{botherref}
\oauthor{\bsnm{Falk}, \binits{T.H.}},
\oauthor{\bsnm{Tam}, \binits{C.}},
\oauthor{\bsnm{Schwellnus}, \binits{H.}},
\oauthor{\bsnm{Chau}, \binits{T.}}:
Grip force variability and its effects on children’s handwriting legibility,
  form, and strokes.
Journal of biomechanical engineering
\textbf{132}(11)
(2010)
\end{botherref}
\endbibitem

\bibitem{lin2017comprehension}
\begin{barticle}
\bauthor{\bsnm{Lin}, \binits{Y.-C.}},
\bauthor{\bsnm{Chao}, \binits{Y.-L.}},
\bauthor{\bsnm{Wu}, \binits{S.-K.}},
\bauthor{\bsnm{Lin}, \binits{H.-H.}},
\bauthor{\bsnm{Hsu}, \binits{C.-H.}},
\bauthor{\bsnm{Hsu}, \binits{H.-M.}},
\bauthor{\bsnm{Kuo}, \binits{L.-C.}}:
\batitle{Comprehension of handwriting development: Pen-grip kinetics in
  handwriting tasks and its relation to fine motor skills among school-age
  children}.
\bjtitle{Australian Occupational Therapy Journal}
\bvolume{64}(\bissue{5}),
\bfpage{369}--\blpage{380}
(\byear{2017})
\end{barticle}
\endbibitem

\bibitem{hug2018surface}
\begin{botherref}
\oauthor{\bsnm{Hug}, \binits{F.}},
\oauthor{\bsnm{Tucker}, \binits{K.}}:
Surface electromyography to study muscle coordination
(2018)
\end{botherref}
\endbibitem

\bibitem{linderman2009recognition}
\begin{barticle}
\bauthor{\bsnm{Linderman}, \binits{M.}},
\bauthor{\bsnm{Lebedev}, \binits{M.A.}},
\bauthor{\bsnm{Erlichman}, \binits{J.S.}}:
\batitle{Recognition of handwriting from electromyography}.
\bjtitle{PLoS One}
\bvolume{4}(\bissue{8}),
\bfpage{6791}
(\byear{2009})
\end{barticle}
\endbibitem

\bibitem{beltran2020multi}
\begin{barticle}
\bauthor{\bsnm{Beltran-Hernandez}, \binits{J.G.}},
\bauthor{\bsnm{Ruiz-Pinales}, \binits{J.}},
\bauthor{\bsnm{Lopez-Rodriguez}, \binits{P.}},
\bauthor{\bsnm{Lopez-Ramirez}, \binits{J.L.}},
\bauthor{\bsnm{Avina-Cervantes}, \binits{J.G.}}:
\batitle{Multi-stroke handwriting character recognition based on semg using
  convolutional-recurrent neural networks}.
\bjtitle{Mathematical Biosciences and Engineering}
\bvolume{17}(\bissue{5}),
\bfpage{5432}--\blpage{5448}
(\byear{2020})
\end{barticle}
\endbibitem

\bibitem{loconsole2019model}
\begin{barticle}
\bauthor{\bsnm{Loconsole}, \binits{C.}},
\bauthor{\bsnm{Cascarano}, \binits{G.D.}},
\bauthor{\bsnm{Brunetti}, \binits{A.}},
\bauthor{\bsnm{Trotta}, \binits{G.F.}},
\bauthor{\bsnm{Losavio}, \binits{G.}},
\bauthor{\bsnm{Bevilacqua}, \binits{V.}},
\bauthor{\bsnm{Di~Sciascio}, \binits{E.}}:
\batitle{A model-free technique based on computer vision and semg for
  classification in parkinson’s disease by using computer-assisted
  handwriting analysis}.
\bjtitle{Pattern Recognition Letters}
\bvolume{121},
\bfpage{28}--\blpage{36}
(\byear{2019})
\end{barticle}
\endbibitem

\bibitem{lansari2003novel}
\begin{bchapter}
\bauthor{\bsnm{Lansari}, \binits{A.}},
\bauthor{\bsnm{Bouslama}, \binits{F.}},
\bauthor{\bsnm{Khasawneh}, \binits{M.}},
\bauthor{\bsnm{Al-Rawi}, \binits{A.}}:
\bctitle{A novel electromyography (emg) based classification approach for
  arabic handwriting}.
In: \bbtitle{Proceedings of the International Joint Conference on Neural
  Networks, 2003.},
vol. \bseriesno{3},
pp. \bfpage{2193}--\blpage{2196}
(\byear{2003}).
\bcomment{IEEE}
\end{bchapter}
\endbibitem

\bibitem{alma2006recognition}
\begin{bchapter}
\bauthor{\bsnm{Alma'adeed}, \binits{S.}}:
\bctitle{Recognition of off-line handwritten arabic words using neural
  network}.
In: \bbtitle{Geometric Modeling and Imaging--New Trends (GMAI'06)},
pp. \bfpage{141}--\blpage{144}
(\byear{2006}).
\bcomment{IEEE}
\end{bchapter}
\endbibitem

\end{thebibliography}


\end{document}